\documentclass[10pt,twocolumn,letterpaper]{article}

\usepackage{wacv}
\usepackage{times}
\usepackage{epsfig}
\usepackage{graphicx}
\usepackage{amsmath}
\usepackage{amsmath, amsthm, amssymb, amsfonts}
\usepackage{amssymb}
\usepackage{booktabs}
\usepackage{tabu,multirow}
\usepackage[mathscr]{euscript}
\usepackage{bbm}
\usepackage{graphicx}
\usepackage{subcaption}
\usepackage{xcolor}
\usepackage{array,multirow}
\usepackage{url}
\usepackage{subfloat}

\newcommand{\gcmpii}{\textit{GazeCapture$\rightarrow$MPIIGaze}}
\newcommand{\gccol}{\textit{GazeCapture$\rightarrow$Columbia}}

\def\wacvPaperID{1091} 

\usepackage[pagebackref=true,breaklinks=true,colorlinks,bookmarks=false]{hyperref}

\wacvfinalcopy

\begin{document}

\title{CUDA-GHR: Controllable Unsupervised Domain Adaptation for Gaze and Head Redirection}

\author{Swati Jindal, Xin Eric Wang\\
University of California, Santa Cruz\\
{\tt\small \{swjindal, xwang366\}@ucsc.edu}
}

\maketitle
\thispagestyle{empty}

\begin{abstract}
    The robustness of gaze and head pose estimation models is highly dependent on the amount of labeled data. Recently, generative modeling has shown excellent results in generating photo-realistic images, which can alleviate the need for annotations. However, adopting such generative models to new domains while maintaining their ability to provide fine-grained control over different image attributes, \eg, gaze and head pose directions, has been a challenging problem. This paper proposes CUDA-GHR, an unsupervised domain adaptation framework that enables fine-grained control over gaze and head pose directions while preserving the appearance-related factors of the person. Our framework simultaneously learns to adapt to new domains and disentangle visual attributes such as appearance, gaze direction, and head orientation by utilizing a label-rich source domain and an unlabeled target domain. Extensive experiments on the benchmarking datasets show that the proposed method can outperform state-of-the-art techniques on both quantitative and qualitative evaluations. Furthermore, we demonstrate the effectiveness of generated image-label pairs in the target domain for pretraining networks for the downstream task of gaze and head pose estimation. The source code and pre-trained models are available at \url{https://github.com/jswati31/cuda-ghr}.
\end{abstract}


\begin{figure}[t!]
    \centering
    \begin{subfigure}{\linewidth}
         \centering
         \includegraphics[width=0.65\textwidth]{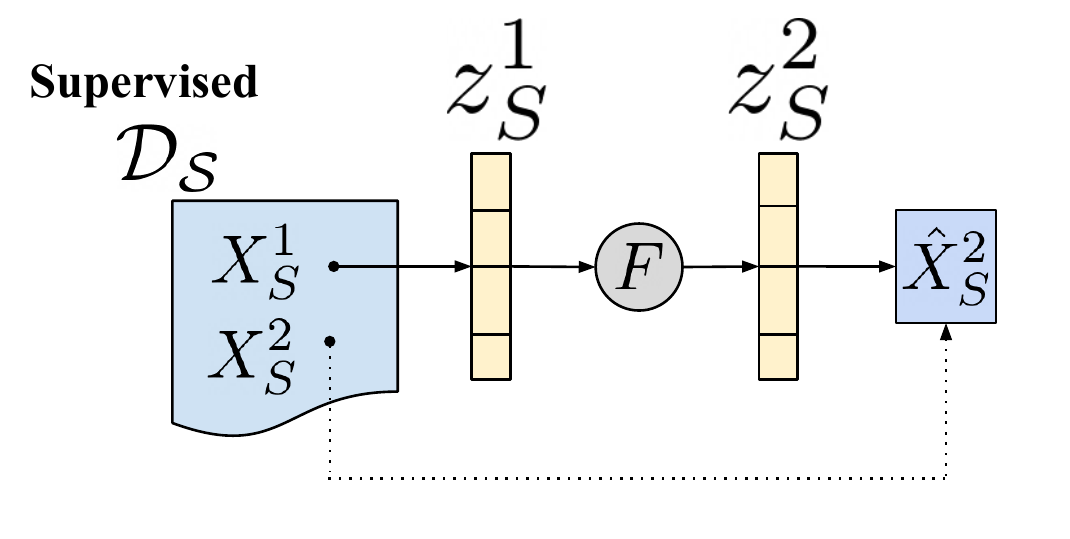}
         \caption{Previous approaches (FAZE~\cite{park2019few}; ST-ED~\cite{zheng2020self})}
     \end{subfigure}
     \hfill
     \begin{subfigure}{\linewidth}
         \centering
         \includegraphics[width=0.7\textwidth]{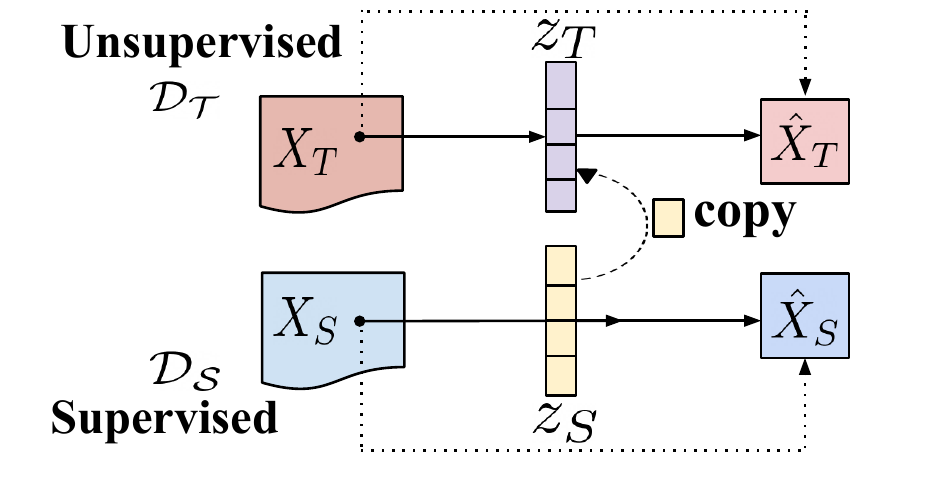}
         \caption{Proposed method}
     \end{subfigure}
    \caption{\textbf{Comparison of existing methods and proposed method}. In Fig (a),  previous approaches~\cite{park2019few, zheng2020self} assume conditional image-to-image translation ($X_S^1 \rightarrow X_S^2$) using a pair of labeled samples from a single domain $\mathcal{D_S}$ and use a transforming function $F$ in the latent space to ensure disentanglement. Here,  $\mathcal{D_S}$ and $\mathcal{D_T}$ represent the source and target domains.  In Fig (b), our method auto-encodes the images $X_S$, $X_T$ from both domains into a common disentangled space using labels only from source, and transfers latent factors via a simple copy operation.}
    \label{fig:comparison}
\end{figure}

\section{Introduction}
Gaze behavior plays a pivotal role in the analysis of non-verbal cues and can provide support to various applications such as virtual reality~\cite{patney2016perceptually, pfeiffer2008towards}, human-computer interaction~\cite{majaranta2014eye,jacob2003eye}, cognition~\cite{buswell1935people,rothkopf2007task}, and social sciences~\cite{ishii2016prediction,oertel2015deciphering}. Recent gaze estimation models rely on learning robust representations requiring a time-consuming and expensive step of collecting a large amount of training data, especially when labels are continuous.  Although various methods~\cite{wood2015rendering,wood2016learning,shrivastava2017learning} have been proposed to circumvent the data need, to generalize in-the-wild real-world scenarios remains a challenge and is an open research problem.

Different gaze redirection methods~\cite{zheng2020self, xia2020controllable, kaur2021subject}
have been explored as an alternate solution for generating more labeled training data using generative adversarial networks (GANs)~\cite{goodfellow2014generative} based frameworks. These generative methods require a pair of labeled images across both source and target domains to learn image-to-image translation; thus, these methods fail to generalize faithfully to new domains. 
Furthermore, various visual attributes are entangled during the generation process and cannot be manipulated independently to provide fine-grained control.
Consequently,  these methods have limited applicability, as in order for the generated data to be useful on downstream tasks, the variability of these visual attributes across the generated data plays a key role in their success. Few works~\cite{9010675,KowalskiECCV2020} on neural image generation attempt to manipulate individual visual attributes in-the-wild real-world scenarios; however, they are constrained by the availability of simulated data with pre-defined labeled attributes. The recent work~\cite{wang2022contrastive} proposes contrastive regression loss and utilizes unsupervised domain adaption to improve gaze estimation performance. 


In this paper, we propose a novel domain adaptation framework for the task of controllable generation of eye gaze and head pose directions in the target domain while not requiring any label information in the target domain. Our method learns to render such control by disentangling explicit factors (\eg, gaze and head orientations) from various implicit factors (\eg, appearance, illumination, shadows, etc.) using a labeled-rich source domain and an unlabeled target domain. Both disentanglement and domain adaptation are performed jointly, thus enabling the transfer of learned knowledge from the source to the target domain. Since we use only unlabeled target-domain data to train our framework, we call it as \textit{unsupervised domain adaptation}~\cite{zou2018unsupervised, toldo2020unsupervised}.

Figure \ref{fig:comparison} illustrates the differences between the proposed method and previous approaches~\cite{park2019few,zheng2020self}. Previous approaches use a pair of labeled samples ($X_S^1, X_S^2$) from the source domain $\mathcal{D_S}$ to learn the conditional image-to-image translation while disentangling visual attributes  using a transforming function $F$. In particular, 
Park \etal~\cite{park2019few} provides control over only explicit factors while Zheng \etal~\cite{zheng2020self} manipulate both explicit and implicit visual attributes. 
In contrast, our method can perform controllable generation  without any input-output paired samples and apply auto-encoding of images $X_S$ and $X_T$ from source $\mathcal{D_S}$ and target $\mathcal{D_T}$ domains into a common disentangled latent space. Concurrently, we adapt the latent representations from the two domains, thereby allowing the transfer of learned knowledge from the labeled source to the unlabeled target domain. Unlike previous approaches, the proposed method is less constrained by label information and can be seamlessly applied to a broader set of datasets/applications.

We train our method on GazeCapture~\cite{krafka2016eye} dataset and demonstrate its efficacy on two target domains: MPIIGaze~\cite{zhang2015appearance} and Columbia~\cite{smith2013gaze} and obtain improved qualitative and quantitative results over state-of-the-art methods~\cite{park2019few, zheng2020self}. Our experimental results exhibit a higher quality in preserving photo-realism of the generated images while faithfully rendering the desired gaze direction and head pose orientation. Overall, our contributions can be summarized as follows: 
\begin{enumerate}
    \item We propose a domain adaptation framework for jointly learning disentanglement and domain adaptation in latent space, using labels only from the source domain. 
    \item Our method utilizes auto-encoding behavior to maintain implicit factors and enable fine-grained control over gaze and head pose directions and outperforms the baseline methods on various evaluation metrics.
    \item We demonstrate the effectiveness of generated redirected images in improving the downstream task performance on gaze and head pose estimation.
\end{enumerate}

\section{Related Work}
\label{relwork}
This section provides a brief overview of the works on learning disentangled representations and gaze redirection methods.

\subsection{Disentangled Representations}
The goal of learning disentangled representations is to model the variability of implicit and explicit factors prevalent in the data generating process \cite{mathieu2016disentangling}. Fully supervised methods~\cite{qian2019make,zakharov2019few,choi2018stargan} exploit the semantic knowledge gained from the available annotations to learn these disentangled representations. On the other hand, unsupervised methods~\cite{higgins2016beta,chen2018isolating} aim to learn the same behavior without relying on any labeled information. However, these methods provide limited flexibility to choose a specific factor of variation and are predominantly focused on a single domain representation learning problems \cite{chen2016infogan}. 

Unsupervised cross-domain disentangled representation learning methods~\cite{lee2018diverse,huang2018multimodal} exploit the advantage of domain-shared and domain-specific attributes in order to provide fine-grained control on the appearance and content of the image. For instance, synthetic data is utilized by a few recent works~\cite{9010675,KowalskiECCV2020} to control various visual attributes while relying on the pre-defined label information associated with the rendered image obtained through a graphics pipeline. On the other hand, Liu \etal~\cite{liu2018detach} provide control over different image attributes using the images from both source and target domains and is trained in a semi-supervised setting. However, their approach only considers categorical labels and thus has limited applicability. In contrast, our method allows controllable manipulation of continuous-valued image attributes (i.e., gaze and head pose) in the cross-domain setting.

\begin{figure*}[h]
    \centering
    \includegraphics[width=\linewidth]{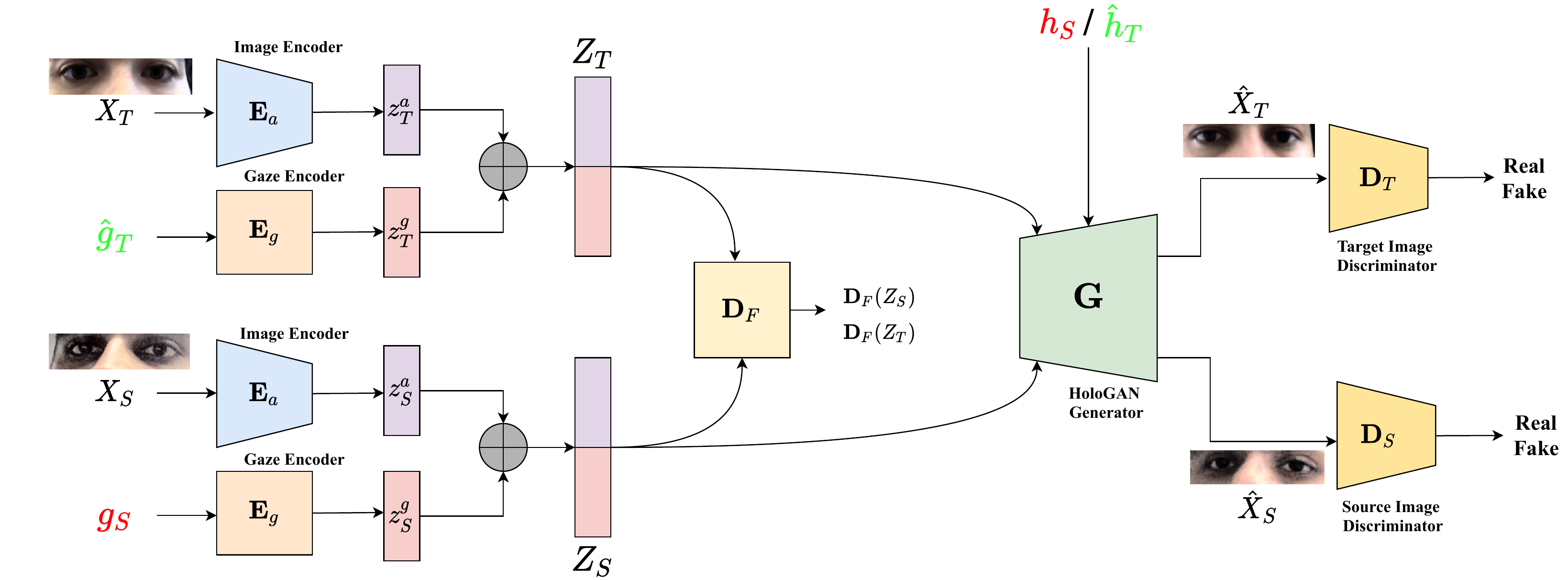}
    \caption{\textbf{Overview of CUDA-GHR.} The framework consists of two encoders $\mathbf{E}_a$ and $\mathbf{E}_g$ shared by both source and target domains. $\mathbf{E}_a$ encodes the target domain image $X_T$ to $z_T^a$, and the source domain image $X_S$ to $z_S^a$ while $\mathbf{E}_g$ encodes the target pseudo gaze label $\hat{g}_T$ and ground-truth source gaze label $g_S$ to $z_T^g$ and $z_S^g$, respectively. The overall image representations are formed as $Z_S$  = $z_S^a$ $\oplus$  $z_S^g$ and $Z_T$  = $z_T^a$ $\oplus$  $z_T^g$ (where, $\oplus$ is concatenate operation). These domain-specific encoded embeddings $Z_T$ and $Z_S$ are passed through a shared generator network $\mathbf{G}$ along with the corresponding head poses (pseudo head pose label $\hat{h}_T$ for the target domain, and ground-truth head pose label $h_S$ for source domain). These embeddings are also passed through a feature domain discriminator $\mathbf{D}_F$. $\mathbf D_T$ and $\mathbf D_S$ represent two domain-specific image discriminators. The whole framework is learned in an end-to-end manner. The labels in \textcolor{red}{red} color are the ground-truth labels while in \textcolor{green}{green} color are the generated pseudo-labels.
    }
    \label{fig:network}
\end{figure*}

\subsection{Gaze Redirection Methods}
Numerous methods have been developed for gaze redirection to attain a large amount of labeled synthetic data for the gaze estimation task. Kononenko \etal~\cite{kononenko2015learning} use random forests to predict the flow field for gaze correction. More recently, several works \cite{ganin2016deepwarp,chen2021coarse,yu2019improving} employ a deep neural network to learn the warping flow field between images along with a correction term. However, these warping-based methods cannot generalize well to large gaze and head pose directions. He \etal~\cite{he2019photo} propose a GAN-based framework that utilizes a cycle consistency loss to learn gaze redirection and generate images with a high resolution. 

In addition, Wood \etal \cite{wood2018gazedirector} uses a graphics pipeline to redirect eye images by fitting morphable models. However, these modeling-based methods make assumptions that do not hold in practice. Mask-based generator networks~\cite{park2019semantic} have also been explored for the task of gaze redirection, though their performance is highly dependent on the accuracy of the segmentation module~\cite{kaur2021subject}. Park \etal \cite{park2019few} utilize a transforming encoder-decoder based network~\cite{hinton2011transforming,worrall2017interpretable} to learn disentanglement in the latent space. Recently, Xia \etal \cite{xia2020controllable} and Zheng \etal \cite{zheng2020self} proposed controllable gaze redirection method using conditional image-to-image translation. However, these methods use a pair of labeled samples during training. As mentioned earlier, our method does not require any paired input-output samples and can be adapted to the target domain without any label data.

\section{Proposed Method}
\label{method}
Our goal is to learn a controller network $\bold{C}$ such that given an input image $X_T$ and subset of explicit factors \{$e_i$\} (\eg, gaze and head pose directions), it generates an image $X_O$ satisfying the attributes described by \{$e_i$\}, i.e.,  $C: (X_T, {e_i}) \rightarrow X_O$. To achieve this, we design a framework that learns to disentangle the latent space and manipulate each explicit factor independently. We start with the assumption that there are three factors of variations: 1) appearance-related, including illumination, shadows, person-specific, etc., which might or might not be explicitly labeled with the dataset, 2) eye gaze direction and 3) head pose orientation. We train our network in an unsupervised domain adaptation setting by utilizing a fully labeled source domain and an unlabeled target domain considering distribution shift across datasets into account.  Recall that we have the gaze and head pose labels only for the source domain. Therefore, we aim to disentangle and control these three factors of variations in the latent space and simultaneously transfer the learned behavior to the unsupervised target domain. We named our framework as CUDA-GHR.

\subsection{Model}
\label{model}
The overall architecture of the CUDA-GHR is shown in Figure \ref{fig:network}. We denote $S$ as the source domain and $T$ as the target domain. Further, following the notations used in \cite{park2019few}, we represent the appearance-related latent factor as $z^a$ and gaze latent factor as $z^g$.

The initial stage of our network consists of two encoders: (a) an image encoder $\mathbf E_a$ encodes the implicit (appearance-related) factors of an image $X_{i}$ and outputs $z_i^a$ such that $i\in \{S, T\}$,
and (b) a separate MLP-based gaze encoder $\mathbf E_g$ encodes the input gaze $g_{i}$ corresponding to the image $X_{i}$ to a latent factor $z_{i}^g$. For the source domain, we use ground-truth gaze label $g_{S}$ as input to $\mathbf E_g$ while for the unlabeled target domain, we input pseudo gaze labels $\hat{g}_{T}$ obtained from a pre-trained task network $\mathcal{T}$ that predicts gaze and head pose of an image. Note that $\mathcal{T}$ is trained only on source domain data. Thus, the overall embedding $Z_{i}$ related to an image $X_{i}$ can be formed by concatenating these two latent factors, i.e.,  $Z_{i} = z_{i}^a \oplus z_{i}^g$ (here $\oplus$ denotes concatenation). Further, $Z_{i}$ and head pose label $h_i$ are given as input to a decoder $\mathbf G$ based on the generator used in HoloGAN~\cite{nguyen2019hologan} as it allows the head pose to be separately controlled without any encoder. This generator $\mathbf G$  decodes the latent $Z_{i}$ and head pose $h_{i}$ to an output image given by $\hat{X_i}$ and is trained in an adversarial manner with the discriminator network $\mathbf D_{i}$. Note again that for labeled source images, we use ground-truth head pose label $h_{S}$ while we take pseudo head pose label $\hat{h}_{T}$ produced by task network $\mathcal{T}$ for unlabeled target domain inputs. In addition, we use a feature domain discriminator $\mathbf D_F$ to ensure that the latent distributions of $Z_S$ and $Z_T$ are similar.

At inference time, the gaze and head pose directions are controlled by passing an image from the target domain $X_T$ through the encoder $\mathbf E_a$ and desired gaze direction  $g$ through $\mathbf E_g$, giving us $\mathbf E_a(X_T)$ and $\mathbf E_{g}(g)$ respectively. These two latent factors are concatenated and passed through the generator $\mathbf G$ along with the desired head pose $h$ to generate an output image $\hat{X}_T^{g, h}$ with gaze $g$ and head pose $h$, i.e.,
\begin{equation}
  \hat{X}_T^{g, h} = \mathbf G(\mathbf E_a(X_T) \oplus \mathbf E_{g}(g), \  h) 
\end{equation}
Likewise, we can also control the individual factor of gaze (or head pose) by providing desired gaze (or head pose) direction and pseudo head  pose (or gaze)  label obtained from $\mathcal{T}$ to generate gaze redirected image given as
\begin{equation}
\begin{aligned}
  \hat{X}_T^{g} &= \mathbf G(\mathbf E_a(X_T) \oplus \mathbf E_{g}(g), \  \hat{h}_{T})
\end{aligned}
\label{gazered}
\end{equation}

and head redirected image given as
\begin{equation}
\begin{aligned}
  \hat{X}_T^{h} &= \mathbf G(\mathbf E_a(X_T) \oplus \mathbf E_{g}(\hat{g}_T), \ h)
\end{aligned}
\label{headred}
\end{equation}

\subsection{Learning Objectives}
The overall objective of our method is to learn a common factorized latent space for both source and target domain such that the individual latent factors can be easily controlled to manipulate target images. To ensure this, we train our framework using multiple objective functions, each of which are explained in detail below.
\paragraph{Reconstruction Loss.} We apply pixel-wise L1 reconstruction loss between the generated image  $\hat X_{i}$ and input image $X_{i}$ to ensure the auto-encoding behavior.
\begin{equation}
    \mathcal{L_R} (\hat X_i, X_i) = \frac{1}{|X_i|} || \hat X_i - X_i ||_1
\end{equation}

\noindent Thus, the total reconstruction loss is defined as
\begin{equation}
    \mathcal{L}_{recon} = \sum_{i\in \{S, T\}}\mathcal{L_R} (\hat X_i, X_i)
\end{equation}
\paragraph{Perceptual Loss.} To ensure that our generated images perceptually match the input images, we apply the perceptual loss~\cite{johnson2016perceptual} which is defined as a mean-square loss between the activations of a pre-trained neural network applied between the generated image $\hat X_{i}$ and input image $X_{i}$. For this, we use VGG-16~\cite{simonyan2014very} network trained on ImageNet~\cite{krizhevsky2012imagenet}.
\begin{equation}
    \mathcal{L_P} (\hat X_i, X_i) = \sum_{l=1}^{4} \frac{1}{|\psi_l (X_i)|} \ || \psi_l (\hat X_i) - \psi_l (X_i) ||_2
\end{equation}
where $\psi$ denotes VGG-16 network. Therefore, our overall perceptual loss becomes
\begin{equation}
    \mathcal{L}_{perc} = \sum_{i\in \{S, T\}}\mathcal{L_P} (\hat X_i, X_i)
\end{equation}
\paragraph{Consistency Loss.} To ensure disentangled behavior between implicit and explicit factors, we apply a consistency loss between the generated image $\hat X_{i}$ and input image $X_{i}$. For this, we use a pre-trained task network $\mathcal{T}$ which predicts the pseudo-labels (gaze and head pose) for an image. The consistency loss consists of two terms: (a) \textit{label consistency loss} is applied between pseudo-labels for input and the generated images to preserve the gaze and head pose information, and (b) \textit{redirection consistency loss} guarantees to preserve the pseudo-labels for redirected images. For (b), we generate gaze and head redirected images using Equation \ref{gazered} and \ref{headred} respectively, by applying gaze and head pose labels from source domain. We enforce the gaze prediction consistency between $\hat{X}_T^g$ and $X_S$, and head pose prediction consistency between $\hat{X}_T^g$ and $\hat{X}_T$, i.e., $\mathcal{T}^g({\hat{X}_T^g})$ = $\mathcal{T}^g({X_S})$ and $\mathcal{T}^h({\hat{X}_T^g})$ = $\mathcal{T}^h({X_T})$. A similar argument holds for the head redirected image $\hat{X}_T^h$, i.e., $\mathcal{T}^g({\hat{X}_T^h})$ = $\mathcal{T}^g({X_T})$ and $\mathcal{T}^h({\hat{X}_T^h})$ = $\mathcal{T}^h({X_S})$. Here, $\mathcal{T}^g$ and $\mathcal{T}^h$ represent the gaze and head pose predicting layers of $\mathcal{T}$. The overall gaze consistency loss will become

\begin{equation}
\begin{aligned}
	\mathcal{L}_{gc} &=
         \underbrace{\mathcal{L}_{a}(\mathcal{T}^g(\hat{X_S}), \mathcal{T}^g({X_S})) + \mathcal{L}_{a}(\mathcal{T}^g(\hat{X_T}), \mathcal{T}^g({X_T}))}_\textit{label consistency loss} \\
        & +\underbrace{\mathcal{L}_{a}(\mathcal{T}^g({\hat{X}_T^g}), \mathcal{T}^g({X_S})) +  \mathcal{L}_{a}(\mathcal{T}^g({\hat{X}_T^h}), \mathcal{T}^g({X_T}))}_\textit{redirection consistency loss}
\end{aligned}
\end{equation} 

Similarly, we can compute the head pose consistency loss $\mathcal{L}_{hc}$ as follows:
\begin{equation}
\begin{aligned}
    \mathcal{L}_{hc} &=
    \underbrace{\mathcal{L}_{a}(\mathcal{T}^h(\hat{X_S}), \mathcal{T}^h({X_S})) + \mathcal{L}_{a}(\mathcal{T}^h(\hat{X_T}), \mathcal{T}^h({X_T}))}_\textit{label consistency loss}\\
     &+ \underbrace{\mathcal{L}_{a}(\mathcal{T}^h({\hat{X}_T^g}), \mathcal{T}^h({X_T})) +  \mathcal{L}_{a}(\mathcal{T}^h({\hat{X}_T^h}), \mathcal{T}^h({X_S}))}_\textit{redirection consistency loss}
\end{aligned}
\end{equation} 

\noindent Here, $\mathcal{L}_{a}$ is defined as:
\begin{equation}
    \mathcal{L}_{a} (\pmb{\hat{u}}, \pmb{u}) = \text{arccos} \left(\dfrac{\pmb{\hat u} \cdot \pmb{u}}{||\pmb{\hat u}|| \cdot ||\pmb{u}||}\right)
\end{equation}
Hence, total consistency loss becomes
\begin{equation}
    \mathcal{L}_{consistency} = \mathcal{L}_{gc} + \mathcal{L}_{hc}
\end{equation}
\paragraph{GAN Loss.} To enforce photo-realistic output from the generator $\mathbf G$, we apply the standard GAN loss~\cite{goodfellow2014generative} to image discriminator $\mathbf D_i$.
\begin{equation}
\begin{aligned}
    \mathcal{L}_{GAN_D} (\mathbf D_i, X_i, \hat{X_i}) &= \text{log} \ \mathbf D_i(X_i) + \text{log}(1 - \mathbf D_i(\hat{X_i})) \\
    \mathcal{L}_{GAN_G} (\mathbf D_i, \hat{X_i}) &= \text{log} \ \mathbf D_i(\hat{X_i})
\end{aligned}
\end{equation}
\noindent The final GAN loss is defined as
\begin{equation}
\begin{aligned}
    \mathcal{L}_{disc} &= \sum_{i\in \{S, T\}}\mathcal{L}_{GAN_D}(\mathbf{D}_i, X_i, \hat{X}_i) \\
    \mathcal{L}_{gen} &= \sum_{i\in \{S, T\}}\mathcal{L}_{GAN_G}(\mathbf{D}_i, \hat{X}_i)
\end{aligned}
\end{equation}
\paragraph{Feature Domain Adversarial Loss.} We employ a latent domain discriminator network $\mathbf{D}_{F}$ and train it using the following domain adversarial loss~\cite{tzeng2017adversarial} to push the distribution of $Z_T$ closer to $Z_S$. 
\begin{equation}
\begin{split}
    \mathcal{L}_{feat} (\mathbf{D}_{F}, Z_T, Z_S) &= \text{log} \ \mathbf{D}_F(Z_S) + \text{log}(1 - \mathbf{D}_F(Z_T))
\end{split}
\end{equation}

\paragraph{Overall Loss.} Altogether, our final loss function for training encoders and generator network is
\begin{equation}
\begin{aligned}
    \mathcal{L}_{overall}& = \lambda_R \mathcal{L}_{recon} + \lambda_P \mathcal{L}_{perc} + \lambda_C \mathcal{L}_{consistency}  \\
   & + \lambda_G \mathcal{L}_{gen} + \lambda_F \mathcal{L}_{feat}
\end{aligned}
\end{equation} 

\section{Experiments}
\label{exp}

\subsection{Datasets}
\label{subsec:data}
\noindent
\textbf{GazeCapture}~\cite{krafka2016eye} is the largest publicly available gaze dataset consisting of around 2M frames taken from unique 1474 subjects. Following the split defined in \cite{krafka2016eye}, we use data from 1274 subjects for training, 50 for validation, and 150 for the test.

\noindent
\textbf{MPIIGaze}
~\cite{zhang2015appearance} is the most challenging dataset for the in-the-wild gaze estimation and includes higher within-subject variations in appearance, for example, illumination, make-up, and facial hair. We use the images from the standard evaluation subset MPIIFaceGaze~\cite{zhang2017s}
provided by MPIIGaze containing 37667 images captured from 15 subjects.

\noindent
\textbf{Columbia}~\cite{smith2013gaze} contains 5880 high-resolution images from 56 subjects and displays larger diversity within participants. The images are collected in a constrained laboratory setting, with limited variations of head pose and gaze directions. 

\begin{table*}[!t]
\caption{\textbf{Quantitative Evaluation.} Comparison of CUDA-GHR with the state-of-the-art methods~\cite{park2019few, zheng2020self}. \gcmpii\ is evaluated on MPIIGaze subsets and \gccol\ is evaluate on Columbia subsets. All errors are in degrees ($^\circ$) except LPIPS, and lower is better.}
\label{tab:quant-mpii}
\centering
\resizebox{\linewidth}{!}
{
  \begin{tabular}{l|l|ccccc|ccccc}
    \hline
    & & \multicolumn{5}{c|}{\textbf{\gcmpii}} & \multicolumn{5}{c}{\textbf{\gccol}}\\
    \hline
    Test Set & Method & LPIPS $\downarrow$   & Gaze  &  Head & $g \rightarrow h$ $\downarrow$ & $h \rightarrow g$ $\downarrow$ & LPIPS $\downarrow$   & Gaze  &  Head & $g \rightarrow h$ $\downarrow$ & $h \rightarrow g$ $\downarrow$ \\
    
     &  &  & Redir. $\downarrow$   &  Redir. $\downarrow$ & & & &  Redir. $\downarrow$   &  Redir. $\downarrow$ &  &   \\

    \hline
    Unseen & 
    
    FAZE  & 0.311   & 6.131   & 6.408   & 6.925   & 4.909 & 0.435   & 9.008   & 6.996   &  6.454  & 4.295    \\
    
   &  ST-ED   & 0.274   & 2.355   & 1.605   & 1.349   & 2.455  & 0.265 & 2.283 & 1.651 & 1.364 & 2.190    \\
   
   &  ST-ED+PS  & 0.266 & 2.864 & 1.576 & 1.472 & 2.346 & 0.266 & 2.117 & 1.437 & \textbf{1.124} & 2.356
    \\
   
   &  \text{CUDA-GHR}  & \textbf{0.261}   & \textbf{2.023}   & \textbf{1.154}   & \textbf{1.161}   & \textbf{1.829} & \textbf{0.255}   & \textbf{1.449}   & \textbf{0.873}   & {1.209}  & \textbf{1.514}    \\
   
    \hline
    Seen & 
    
    FAZE  & 0.382   & 5.778   & 6.899   & 5.311   & 5.172 & 0.486   & 10.368   & 7.231   &  7.302  & 4.788   \\
    
    & \text{ST-ED}  & 0.315   & 2.405   & 1.669   & 1.209   & 2.341 & 0.319   & 2.484   & 1.616   & 1.343   & 2.456    \\
    
   &  ST-ED+PS  & 0.288 & 2.269 & 1.888  & 1.179 & 2.229 & 0.299 & 2.071 & 1.536 & 1.088  & 2.330   \\
    
    & \text{CUDA-GHR}  & \textbf{0.278}   & \textbf{1.905}   & \textbf{0.979}   & \textbf{0.761}   & \textbf{1.236}  & \textbf{0.282}   & \textbf{1.328}   & \textbf{0.831}   & \textbf{0.646}   & \textbf{0.996}   \\
    
    \hline
    All & 
    
    FAZE  & 0.370   & 5.840   & 6.828   & 5.613   & 5.123  & 0.481   & 10.214   & 7.226   &  7.214  & 4.737    \\
    
    & \text{ST-ED}  & 0.307   & 2.392   & 1.660   & 1.232   & 2.359   & 0.314   & 2.473   & 1.618   & 1.350   & 2.435    \\
    
    & \text{CUDA-GHR}   & \textbf{0.275}   & \textbf{1.922}   & \textbf{1.012}   & \textbf{0.844}   & \textbf{1.341} & \textbf{0.279}  & \textbf{1.337}   & \textbf{0.832}   & \textbf{0.707}  & \textbf{1.045}    \\
    \hline

  \end{tabular}
  }

\end{table*}

\subsection{Implementation Details}
The architecture of the encoder $\mathbf  E_a$ is DenseNet-based blocks as used in Park \etal \cite{park2019few} and the decoder network $\mathbf  G$ is HoloGAN based generator~\cite{nguyen2019hologan}. The gaze encoder $\mathbf  E_g$ consists of four MLP layers with hidden dimensions equal to the input dimension and output dimension is set to 8.  The task network $\mathcal{T}$ is a ResNet-50~\cite{he2016deep} based model trained on GazeCapture~\cite{krafka2016eye} training subset and gives 4-D output, two angles for each gaze and head direction. The two image discriminators $\mathbf  D_S$ and $\mathbf  D_T$ share a similar PatchGAN~\cite{isola2017image} based architecture. The domain discriminator $\mathbf  D_{F}$ consists of four MLP layers. Note that $\mathcal{T}$ remains fixed during training of our whole pipeline. More implementation details can be found in the supplementary materials.

All the datasets are pre-processed by a data normalization algorithm as described in Zhang \etal \cite{zhang2018revisiting}. Our input is a single image containing both eyes and is of size 256 $\times$ 64. We use a data processing pipeline as employed in Park \etal \cite{park2019few} to extract the eye image strip. The inputs gaze $g$ and head pose $h$ are 2-D pitch and yaw angles.  We train our framework in two settings: \gcmpii, trained with  GazeCapture as source domain and MPIIGaze as target domain, and \gccol\  is trained with Columbia as the target domain. For GazeCapture, we use the training subset from the data split as labeled source domain data. From MPIIGaze and Columbia, we respectively choose the first 11 and 50 subjects as unlabeled target domain data for training. We call them as \textbf{`Seen'} subjects as our network sees them during training while remaining users fall into \textbf{`Unseen'} category. We evaluate our method on three test subsets: `Unseen', `Seen' and `All'. `All' includes both `Seen' and `Unseen' participants data. 

\paragraph{Hyper-parameters.} We use a batch size of 10 for both \gcmpii \ and \gccol \ and are trained for 200K and 81K iterations, respectively. All network modules are optimized through Adam~\cite{kingma2014adam} optimizer with a weight decay coefficient of $10^{-4}$. The initial learning rate is set to $0.0005$ which is decayed by a factor of $0.8$ after approximately 34K iterations. For \gcmpii , we restart the learning rate scheduler after around 160K iterations for better convergence. The weights of the objective function are set as $\lambda_R$ = 200, $\lambda_P$ = 10, $\lambda_C$ = 10, $\lambda_G$ = 5 and $\lambda_F$ = 5.

\subsection{Evaluation Metrics}
We evaluate our framework using three evaluation metrics as previously adopted by ~\cite{zheng2020self}: perceptual similarity, redirection errors, and disentanglement errors.\\

\noindent
\textbf{Learned Perceptual Image Patch Similarity (LPIPS)}~\cite{zhang2018unreasonable} is used to measure the pairwise image similarity by calculating the distance in AlexNet~\cite{krizhevsky2014one} feature space.

\noindent
\textbf{Redirection Errors} are computed as angular errors between the estimated direction obtained from our task network $\mathcal{T}$ and the desired direction. It measures the accomplishment of the explicit factors, i.e., gaze and head directions in the image output.

\noindent
\textbf{Disentanglement Error} measures the disentanglement of explicit factors like gaze and head pose. We evaluate $g\rightarrow h$, the effect of change in gaze direction on the head pose, and vice versa ($h\rightarrow g$). To compute $g\rightarrow h$, we first calculate the joint probability distribution function of the gaze direction values from the source domain and sample random gaze labels. We apply this gaze direction to the input image while keeping the head pose unchanged and measure the angular error between head pose predictions from task network $\mathcal{T}$ of the redirected image and the original reconstructed image. Similarly, we compute $h\rightarrow g$ by sampling random head pose orientations from the source labeled data.

\subsection{Comparison to the state-of-the-art}
\label{sub:comparison}
We adopt FAZE~\cite{park2019few} and ST-ED~\cite{zheng2020self}
as our baseline methods. Both FAZE and ST-ED are based on transforming encoder-decoder architecture~\cite{hinton2011transforming,worrall2017interpretable} and apply known differences in gaze and head rotations to the embedding space for translating the input image to a redirected output image. FAZE inputs an image containing both eyes, which is the same as our method, thus necessary to compare. 
We use original implementation\footnote{\url{https://github.com/NVlabs/few_shot_gaze}} and trained models provided by the FAZE authors for comparison. 
In addition, we re-train the ST-ED network on images containing both eyes 
for a fair comparison. FAZE learns to control only explicit factors (gaze and head pose orientations) while ST-ED controls implicit factors too. Note that for the ST-ED baseline, we compare only by altering explicit factors. Furthermore, we also compare CUDA-GHR to baseline ST-ED+PS which is trained with source data GazeCapture and using pseudo-labels for target dataset (MPIIGaze or Columbia). The pseudo-labels are obtained in same manner as of CUDA-GHR. For more details, please refer to the supplementary materials.

\paragraph{Quantitative Evaluation.}
Table \ref{tab:quant-mpii} summarizes the quantitative evaluation of both our experiments \gcmpii{} and \gccol{}. The left half of Table \ref{tab:quant-mpii} shows evaluation on MPIIGaze test subsets \{`Seen', `Unseen', `All'\}, and we observe that our method outperforms the baselines (even ST-ED+PS) on all the evaluation metrics for each test subset. We get lower LPIPS (even on ‘Unseen’ users), indicating the generation of better quality images while achieving the desired gaze and head directions attested by lower gaze and head redirection errors. We also obtain better disentanglement errors exhibiting that our method successfully controls each explicit factor individually.  The improved performance on `Unseen' users shows the superiority and generalizability of our method over baselines. We also notice improvements over ST-ED+PS baseline, exhibiting that domain adaptation is essential to achieve better performance.

We show evaluation of \gccol \ experiment on right half of Table \ref{tab:quant-mpii}. Note that due to the small size of the Columbia dataset, we initialize the model for this experiment with the previously trained weights on \gcmpii{} for better convergence. Recall that we do not use any labels from the target domain dataset in any experiment. As shown in Table \ref{tab:quant-mpii}, our method is consistently better than other baselines on all evaluation metrics, showing the generalizability of our framework on different domains and thus, can be adapted to new datasets without the requirement of any labels.

\begin{figure}[t!]
    \centering
    \begin{subfigure}[h]{\linewidth}
        \includegraphics[width=\linewidth, height=5cm]{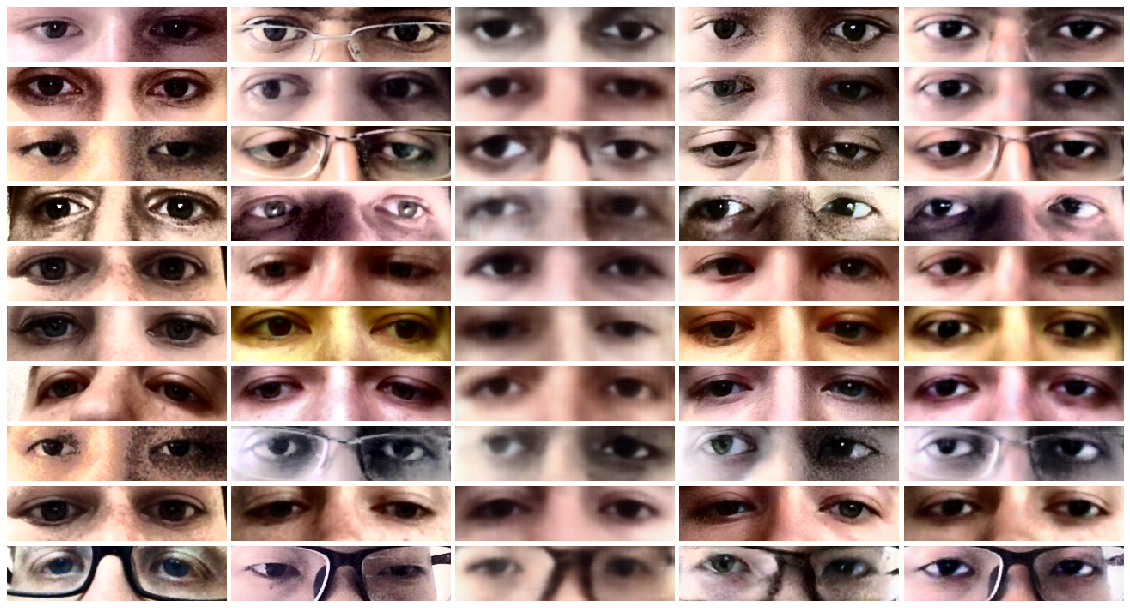}
        \vspace{-0.7cm}
        \caption*{\scriptsize{Gaze Source \hspace{0.5cm} Input Image  \hspace{0.5cm} FAZE  \hspace{0.85cm} ST-ED  \hspace{0.7cm}  CUDA-GHR}}
        \caption{Gaze Redirected images}
        \label{fig:gazeswapped}
    \end{subfigure}
	\hfill
	\begin{subfigure}[h]{\linewidth}
    \centering
        \includegraphics[width=1\linewidth, height=5cm]{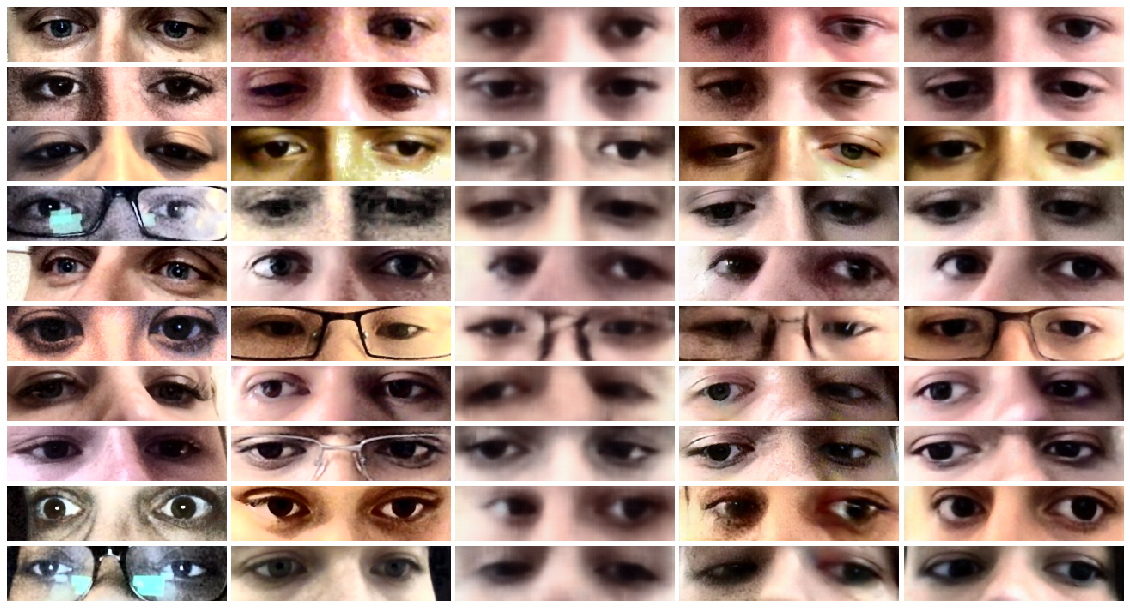}
         \vspace{-0.7cm}
        \caption*{\scriptsize{Head Source \hspace{0.5cm} Input Image  \hspace{0.5cm} FAZE  \hspace{0.85cm} ST-ED  \hspace{0.7cm}  CUDA-GHR}}
        \caption{Head Redirected images}
        \label{fig:headswapped}
    \end{subfigure}
       \caption{\textbf{Qualitative Evaluation: }Comparison of the generated images from CUDA-GHR (\gcmpii) with the baseline methods FAZE~\cite{park2019few} and ST-ED~\cite{zheng2020self}. The quality of gaze redirected images is depicted in \ref{fig:gazeswapped}, while head redirected images are shown in \ref{fig:headswapped}. The first column represents the gaze/head pose source image from which gaze/head pose information is used to redirect. The second column shows the input image from the target domain. Our method (column 5) produces better quality images and preserves the implicit factors than the baseline methods (columns 3 and 4). Best viewed in color.} \label{fig:quality}
\end{figure}

\paragraph{Qualitative Evaluation.}
We also report the qualitative comparison of generated images in Figure \ref{fig:quality} using a model trained with \gcmpii. The results are shown on MPIIGaze dataset images which is the target domain dataset in this setting. As can be seen, our method produces better quality images while preserving the appearance information (\eg, skin color, eye shape) and faithfully manipulating the gaze and head pose directions when compared with FAZE~\cite{park2019few} and ST-ED~\cite{zheng2020self}. It is also worth noting that our method generates higher-quality images for people with glasses, \eg, row 3 in Figure \ref{fig:gazeswapped} and row 6 in Figure \ref{fig:headswapped}. These results are consistent with our findings in quantitative evaluation, thus showing that our method is more faithful in reproducing the desired gaze and head pose directions. Additional results are provided in the supplementary materials.

\subsection{Ablation Study}
\label{ablation}
To understand the role of individual components of the objective function, we provide following ablation study.
In Table \ref{tab:ablationtable}, we compare against the ablations of individual loss terms. The ablation on the perceptual loss is shown in the first row ($\lambda_P=0$). The second row ($\lambda_C=0$) represents when consistency loss is set to zero, while the third row ($\lambda_F=0$) shows results when feature domain adversarial loss is not enforced during training. The fourth and fifth row shows an ablation on reconstruction ($\lambda_R = 0$) and GAN ($\lambda_G = 0$) loss, respectively.  As can be seen, all of these loss terms are critical for the improvements in performance. We see a substantial improvement with the addition of $\mathcal{L}_{consistency}$. The ablation study is performed for \textit{GazeCapture $\rightarrow$ MPIIGaze} on the `Seen' subset of MPIIGaze.

\begin{table}[t]
\caption{\textbf{Ablation Study: }An ablation study on different loss terms for \textit{GazeCapture $\rightarrow$ MPIIGaze} on MPIIGaze `Seen' subset. All errors are in degrees ($^\circ$) except LPIPS, and lower is better.}
\label{tab:ablationtable}
\centering
\resizebox{\columnwidth}{!}
{
  \begin{tabular}{l|ccccc}
    \hline
     Ablation term & LPIPS $\downarrow$   & Gaze  &  Head & $g \rightarrow h$ $\downarrow$ & $h \rightarrow g$ $\downarrow$  \\
    
     &  & Redir. $\downarrow$   &  Redir. $\downarrow$ &  &   \\
    \hline
    $\lambda_P = 0$  &  0.307  & 6.450  & 0.922 &  0.655  &  1.334 \\

   $ \lambda_C = 0$  &  0.326  &  15.183 &  3.412  &  \textbf{0.106} & 11.616 \\
    
    $\lambda_F = 0$ &  0.281  &  4.791  &  \textbf{0.787}  &   0.636 &  \textbf{0.826} \\
    
   $ \lambda_R = 0$ & 0.304 & 4.958 & 0.911 & 0.463 & 0.876   \\
    
   $ \lambda_G = 0$ & 0.309  & 11.130 & 0.942 & 0.355 & 0.868 \\
    
    Ours & \textbf{0.278}   &  \textbf{1.905}  &  0.979  &  0.761 &  1.236 \\
    \hline
  \end{tabular}
  }
\end{table}

\subsection{Controllability}
\begin{figure}[t]
    \centering
    \begin{subfigure}{\linewidth}
         \centering
         \includegraphics[width=\columnwidth, height=3cm]{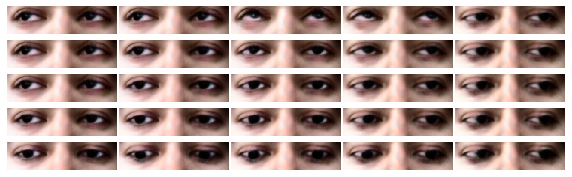}
         \caption{Gaze redirected images with (pitch, yaw) $\in$ [$-30^\circ$, $30^\circ$]}
     \end{subfigure}
     \hfill
     \begin{subfigure}{\linewidth}
         \centering
         \includegraphics[width=\columnwidth, height=3cm]{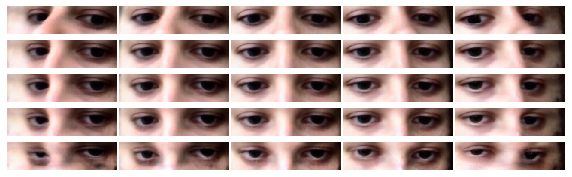}
         \caption{Head redirected images with (pitch, yaw) $\in$ [$-30^\circ$, $30^\circ$]}
     \end{subfigure}
    \caption{\textbf{Controllable Generation: }Illustration of controllable gaze and head redirection showing the effectiveness of disentanglement between various explicit factors.}
    \label{fig:gazecontrol}
\end{figure}

Figure \ref{fig:gazecontrol} shows the effectiveness of our method in controlling the gaze and head pose directions. We vary pitch and yaw angles from $-30^\circ$ to $+30^\circ$ for gaze and head redirections.  We can see that our method faithfully renders the desired gaze direction (or head pose orientation) while retaining the head pose (or gaze direction), therefore, exhibiting the efficacy of disentanglement. Furthermore, note that the range of yaw and pitch angles [$-30^\circ$, $30^\circ$] is out-of-label distribution of source dataset (GazeCapture), showing the extrapolation capability of CUDA-GHR in the generation process.

\subsection{Evaluation of Downstream Tasks}
We also demonstrate the utility of generated images from our framework in improving the performance of the downstream gaze and head pose estimation task. For this, we conduct experiments for cross-subject estimation on both MPIIGaze and Columbia datasets. The main goal of this experiment is to show that the generated ``free" labeled data from our framework can be used to obtain a good pretrained model to further fine-tune on cross-subject estimation task. We compare it against three initializations: random,  ImageNet~\cite{deng2009imagenet}, and pretrained model obtained using ST-ED~\cite{zheng2020self} generated images.

We generate around 10K samples per user from MPIIGaze dataset using \gcmpii{} trained generator and train a ResNet-50~\cite{he2016deep} network (initialized with ImageNet pre-trained weights) with batch normalization~\cite{ioffe2015batch} replaced by instance normalization~\cite{ulyanov2016instance} layers. Afterward, we fine-tune this network on MPIIGaze dataset using leave-one-subject-out cross-validation for both gaze and head pose estimation and report the mean angular error. A similar method is followed for ST-ED generated images. We compare the errors obtained from four initialization methods: random, ImageNet, ST-ED, and CUDA-GHR. Analogously, we train gaze and head pose estimation models on generated images for Columbia data subjects ($\sim$1.6K samples each) using \gccol \ model and fine-tune on Columbia dataset using 4-fold cross-validation. The comparison of different initialization methods on two datasets is shown in Table \ref{tab:crosssubject}.

It can be seen that the model trained with CUDA-GHR gives around $7\%$  and $4\%$ relative improvements over ST-ED initialization on Columbia and MPIIGaze, respectively, for the head pose estimation task. We also show results for the gaze estimation task in Table \ref{tab:crosssubject} giving a relative improvement of around $5.5\%$ on the Columbia dataset while performing similar to the ST-ED baseline on MPIIGaze. We hypothesize that this is because the gaze and head pose label distribution of GazeCapture is closer to MPIIGaze distribution than Columbia~\cite{cheng2021appearance} and thus, performs closely for both ST-ED and CUDA-GHR. This indicates that domain adaptation is more advantageous for the Columbia dataset. Hence, it shows the effectiveness of our method over baselines when performing domain adaptation across datasets with significant distribution shifts. 

\begin{table}[t]
\caption{\textbf{Downstream Task Evaluation: }Comparison of mean angular errors (\textit{mean $\pm$ std} in degrees) for various initialization methods on gaze and head pose estimation task. Lower is better.}
\label{tab:crosssubject}
\centering
\resizebox{\columnwidth}{!}
{
  \begin{tabular}{c|cc|cc}
    \hline
    {\textbf{Initialization}} & \multicolumn{2}{c|}{{\textbf{Head Pose}}} & \multicolumn{2}{c}{{\textbf{Gaze}}} \\
     \textbf{Method} & \multicolumn{2}{c|}{{\textbf{Estimation Errors$\downarrow$}}} & \multicolumn{2}{c}{{\textbf{Estimation Errors$\downarrow$}}} \\
    \hline
    &  Columbia & MPIIGaze  & Columbia & MPIIGaze  \\
    \hline
    \text{Random}   &  6.8 $\pm$ 1.2  &  6.7 $\pm$ 0.7  &   6.7 $\pm$ 0.7 &  6.7 $\pm$ 1.3  \\
    
   \text{ImageNet}  &  5.9 $\pm$ 1.3   &  5.7 $\pm$ 2.8  & 5.5 $\pm$ 0.1   & 5.7 $\pm$ 1.4    \\
 
   \text{ST-ED}    &  5.7 $\pm$ 1.1   & 5.1 $\pm$ 2.4   & 5.4 $\pm$ 0.4  & \textbf{5.5} $\pm$ \textbf{1.3}   \\
   
   \text{CUDA-GHR}     &  \textbf{5.3} $\pm$ \textbf{1.1}  & \textbf{4.9} $\pm$ \textbf{2.5}  & \textbf{5.1} $\pm$ \textbf{0.4} & \textbf{5.5} $\pm$ \textbf{1.4}     \\
    \hline

  \end{tabular}
  }
\end{table}

\section{Conclusion}
We present an unsupervised domain adaptation framework trained using cross-domain datasets for gaze and head redirection tasks. The proposed method takes advantage of both supervised source domain and unsupervised target domain to learn the disentangled factors of variations. Experimental results demonstrate the effectiveness of our model in generating photo-realistic images in multiple domains while truly adapting the desired gaze direction and head pose orientation. Because of removing the requirement of annotations in the target domain, the applicability of our work increases for new datasets where manual annotations are hard to collect. Our framework is relevant to various applications such as video conferencing, photo correction, and movie editing for redirecting gaze to establish eye contact with the viewer. It can also be extended to improve performances on the downstream task of gaze and head pose estimation.

{\small
\bibliographystyle{ieee_fullname}
\bibliography{egbib}

\begin{thebibliography}{10}\itemsep=-1pt

\bibitem{buswell1935people}
Guy~Thomas Buswell.
\newblock How people look at pictures: a study of the psychology and perception
  in art.
\newblock 1935.

\bibitem{chen2021coarse}
Jingjing Chen, Jichao Zhang, Enver Sangineto, Tao Chen, Jiayuan Fan, and Nicu
  Sebe.
\newblock Coarse-to-fine gaze redirection with numerical and pictorial
  guidance.
\newblock In {\em Proceedings of the IEEE/CVF Winter Conference on Applications
  of Computer Vision}, pages 3665--3674, 2021.

\bibitem{chen2018isolating}
Ricky~TQ Chen, Xuechen Li, Roger Grosse, and David Duvenaud.
\newblock Isolating sources of disentanglement in variational autoencoders.
\newblock {\em arXiv preprint arXiv:1802.04942}, 2018.

\bibitem{chen2016infogan}
Xi Chen, Yan Duan, Rein Houthooft, John Schulman, Ilya Sutskever, and Pieter
  Abbeel.
\newblock Infogan: Interpretable representation learning by information
  maximizing generative adversarial nets.
\newblock {\em arXiv preprint arXiv:1606.03657}, 2016.

\bibitem{cheng2021appearance}
Yihua Cheng, Haofei Wang, Yiwei Bao, and Feng Lu.
\newblock Appearance-based gaze estimation with deep learning: A review and
  benchmark.
\newblock {\em arXiv preprint arXiv:2104.12668}, 2021.

\bibitem{choi2018stargan}
Yunjey Choi, Minje Choi, Munyoung Kim, Jung-Woo Ha, Sunghun Kim, and Jaegul
  Choo.
\newblock Stargan: Unified generative adversarial networks for multi-domain
  image-to-image translation.
\newblock In {\em Proceedings of the IEEE conference on computer vision and
  pattern recognition}, pages 8789--8797, 2018.

\bibitem{deng2009imagenet}
Jia Deng, Wei Dong, Richard Socher, Li-Jia Li, Kai Li, and Li Fei-Fei.
\newblock Imagenet: A large-scale hierarchical image database.
\newblock In {\em 2009 IEEE conference on computer vision and pattern
  recognition}, pages 248--255. Ieee, 2009.

\bibitem{deng2018cascade}
Jiankang Deng, Yuxiang Zhou, Shiyang Cheng, and Stefanos Zaferiou.
\newblock Cascade multi-view hourglass model for robust 3d face alignment.
\newblock In {\em 2018 13th IEEE International Conference on Automatic Face \&
  Gesture Recognition (FG 2018)}, pages 399--403. IEEE, 2018.

\bibitem{ganin2016deepwarp}
Yaroslav Ganin, Daniil Kononenko, Diana Sungatullina, and Victor Lempitsky.
\newblock Deepwarp: Photorealistic image resynthesis for gaze manipulation.
\newblock In {\em European conference on computer vision}, pages 311--326.
  Springer, 2016.

\bibitem{goodfellow2014generative}
Ian~J Goodfellow, Jean Pouget-Abadie, Mehdi Mirza, Bing Xu, David Warde-Farley,
  Sherjil Ozair, Aaron Courville, and Yoshua Bengio.
\newblock Generative adversarial networks.
\newblock {\em arXiv preprint arXiv:1406.2661}, 2014.

\bibitem{he2016deep}
Kaiming He, Xiangyu Zhang, Shaoqing Ren, and Jian Sun.
\newblock Deep residual learning for image recognition.
\newblock In {\em Proceedings of the IEEE conference on computer vision and
  pattern recognition}, pages 770--778, 2016.

\bibitem{he2019photo}
Zhe He, Adrian Spurr, Xucong Zhang, and Otmar Hilliges.
\newblock Photo-realistic monocular gaze redirection using generative
  adversarial networks.
\newblock In {\em Proceedings of the IEEE/CVF International Conference on
  Computer Vision}, pages 6932--6941, 2019.

\bibitem{higgins2016beta}
Irina Higgins, Loic Matthey, Arka Pal, Christopher Burgess, Xavier Glorot,
  Matthew Botvinick, Shakir Mohamed, and Alexander Lerchner.
\newblock beta-vae: Learning basic visual concepts with a constrained
  variational framework.
\newblock 2016.

\bibitem{hinton2011transforming}
Geoffrey~E Hinton, Alex Krizhevsky, and Sida~D Wang.
\newblock Transforming auto-encoders.
\newblock In {\em International conference on artificial neural networks},
  pages 44--51. Springer, 2011.

\bibitem{hu2017finding}
Peiyun Hu and Deva Ramanan.
\newblock Finding tiny faces.
\newblock In {\em Proceedings of the IEEE conference on computer vision and
  pattern recognition}, pages 951--959, 2017.

\bibitem{huang2017densely}
Gao Huang, Zhuang Liu, Laurens Van Der~Maaten, and Kilian~Q Weinberger.
\newblock Densely connected convolutional networks.
\newblock In {\em Proceedings of the IEEE conference on computer vision and
  pattern recognition}, pages 4700--4708, 2017.

\bibitem{huang2017arbitrary}
Xun Huang and Serge Belongie.
\newblock Arbitrary style transfer in real-time with adaptive instance
  normalization.
\newblock In {\em Proceedings of the IEEE International Conference on Computer
  Vision}, pages 1501--1510, 2017.

\bibitem{huang2018multimodal}
Xun Huang, Ming-Yu Liu, Serge Belongie, and Jan Kautz.
\newblock Multimodal unsupervised image-to-image translation.
\newblock In {\em Proceedings of the European conference on computer vision
  (ECCV)}, pages 172--189, 2018.

\bibitem{huber2016multiresolution}
Patrik Huber, Guosheng Hu, Rafael Tena, Pouria Mortazavian, P Koppen, William~J
  Christmas, Matthias Ratsch, and Josef Kittler.
\newblock A multiresolution 3d morphable face model and fitting framework.
\newblock In {\em Proceedings of the 11th International Joint Conference on
  Computer Vision, Imaging and Computer Graphics Theory and Applications},
  2016.

\bibitem{ioffe2015batch}
Sergey Ioffe and Christian Szegedy.
\newblock Batch normalization: Accelerating deep network training by reducing
  internal covariate shift.
\newblock In {\em International conference on machine learning}, pages
  448--456. PMLR, 2015.

\bibitem{ishii2016prediction}
Ryo Ishii, Kazuhiro Otsuka, Shiro Kumano, and Junji Yamato.
\newblock Prediction of who will be the next speaker and when using gaze
  behavior in multiparty meetings.
\newblock {\em ACM Transactions on Interactive Intelligent Systems (TIIS)},
  6(1):1--31, 2016.

\bibitem{isola2017image}
Phillip Isola, Jun-Yan Zhu, Tinghui Zhou, and Alexei~A Efros.
\newblock Image-to-image translation with conditional adversarial networks.
\newblock In {\em Proceedings of the IEEE conference on computer vision and
  pattern recognition}, pages 1125--1134, 2017.

\bibitem{jacob2003eye}
Robert~JK Jacob and Keith~S Karn.
\newblock Eye tracking in human-computer interaction and usability research:
  Ready to deliver the promises.
\newblock In {\em The mind's eye}, pages 573--605. Elsevier, 2003.

\bibitem{johnson2016perceptual}
Justin Johnson, Alexandre Alahi, and Li Fei-Fei.
\newblock Perceptual losses for real-time style transfer and super-resolution.
\newblock In {\em European conference on computer vision}, pages 694--711.
  Springer, 2016.

\bibitem{kaur2021subject}
Harsimran Kaur and Roberto Manduchi.
\newblock Subject guided eye image synthesis with application to gaze
  redirection.
\newblock In {\em Proceedings of the IEEE/CVF Winter Conference on Applications
  of Computer Vision}, pages 11--20, 2021.

\bibitem{kingma2014adam}
Diederik~P Kingma and Jimmy Ba.
\newblock Adam: A method for stochastic optimization.
\newblock {\em arXiv preprint arXiv:1412.6980}, 2014.

\bibitem{kononenko2015learning}
Daniil Kononenko and Victor Lempitsky.
\newblock Learning to look up: Realtime monocular gaze correction using machine
  learning.
\newblock In {\em Proceedings of the IEEE Conference on Computer Vision and
  Pattern Recognition}, pages 4667--4675, 2015.

\bibitem{KowalskiECCV2020}
Marek Kowalski, Stephan~J. Garbin, Virginia Estellers, Tadas Baltrušaitis,
  Matthew Johnson, and Jamie Shotton.
\newblock Config: Controllable neural face image generation.
\newblock In {\em European Conference on Computer Vision (ECCV)}, 2020.

\bibitem{krafka2016eye}
Kyle Krafka, Aditya Khosla, Petr Kellnhofer, Harini Kannan, Suchendra
  Bhandarkar, Wojciech Matusik, and Antonio Torralba.
\newblock Eye tracking for everyone.
\newblock In {\em Proceedings of the IEEE conference on computer vision and
  pattern recognition}, pages 2176--2184, 2016.

\bibitem{krizhevsky2014one}
Alex Krizhevsky.
\newblock One weird trick for parallelizing convolutional neural networks.
\newblock {\em arXiv preprint arXiv:1404.5997}, 2014.

\bibitem{krizhevsky2012imagenet}
Alex Krizhevsky, Ilya Sutskever, and Geoffrey~E Hinton.
\newblock Imagenet classification with deep convolutional neural networks.
\newblock {\em Advances in neural information processing systems},
  25:1097--1105, 2012.

\bibitem{lee2018diverse}
Hsin-Ying Lee, Hung-Yu Tseng, Jia-Bin Huang, Maneesh Singh, and Ming-Hsuan
  Yang.
\newblock Diverse image-to-image translation via disentangled representations.
\newblock In {\em Proceedings of the European conference on computer vision
  (ECCV)}, pages 35--51, 2018.

\bibitem{liu2018detach}
Yen-Cheng Liu, Yu-Ying Yeh, Tzu-Chien Fu, Sheng-De Wang, Wei-Chen Chiu, and
  Yu-Chiang~Frank Wang.
\newblock Detach and adapt: Learning cross-domain disentangled deep
  representation.
\newblock In {\em Proceedings of the IEEE Conference on Computer Vision and
  Pattern Recognition}, pages 8867--8876, 2018.

\bibitem{majaranta2014eye}
P{\"a}ivi Majaranta and Andreas Bulling.
\newblock Eye tracking and eye-based human--computer interaction.
\newblock In {\em Advances in physiological computing}, pages 39--65. Springer,
  2014.

\bibitem{mathieu2016disentangling}
Michael Mathieu, Junbo Zhao, Pablo Sprechmann, Aditya Ramesh, and Yann LeCun.
\newblock Disentangling factors of variation in deep representations using
  adversarial training.
\newblock {\em arXiv preprint arXiv:1611.03383}, 2016.

\bibitem{nguyen2019hologan}
Thu Nguyen-Phuoc, Chuan Li, Lucas Theis, Christian Richardt, and Yong-Liang
  Yang.
\newblock Hologan: Unsupervised learning of 3d representations from natural
  images.
\newblock In {\em Proceedings of the IEEE/CVF International Conference on
  Computer Vision}, pages 7588--7597, 2019.

\bibitem{oertel2015deciphering}
Catharine Oertel, Kenneth~A Funes~Mora, Joakim Gustafson, and Jean-Marc Odobez.
\newblock Deciphering the silent participant: On the use of audio-visual cues
  for the classification of listener categories in group discussions.
\newblock In {\em Proceedings of the 2015 ACM on International Conference on
  Multimodal Interaction}, pages 107--114, 2015.

\bibitem{park2019few}
Seonwook Park, Shalini~De Mello, Pavlo Molchanov, Umar Iqbal, Otmar Hilliges,
  and Jan Kautz.
\newblock Few-shot adaptive gaze estimation.
\newblock In {\em Proceedings of the IEEE/CVF International Conference on
  Computer Vision}, pages 9368--9377, 2019.

\bibitem{park2019semantic}
Taesung Park, Ming-Yu Liu, Ting-Chun Wang, and Jun-Yan Zhu.
\newblock Semantic image synthesis with spatially-adaptive normalization.
\newblock In {\em Proceedings of the IEEE/CVF Conference on Computer Vision and
  Pattern Recognition}, pages 2337--2346, 2019.

\bibitem{patney2016perceptually}
Anjul Patney, Joohwan Kim, Marco Salvi, Anton Kaplanyan, Chris Wyman, Nir
  Benty, Aaron Lefohn, and David Luebke.
\newblock Perceptually-based foveated virtual reality.
\newblock In {\em ACM SIGGRAPH 2016 Emerging Technologies}, pages 1--2. 2016.

\bibitem{pfeiffer2008towards}
Thies Pfeiffer.
\newblock Towards gaze interaction in immersive virtual reality: Evaluation of
  a monocular eye tracking set-up.
\newblock In {\em Virtuelle und Erweiterte Realit{\"a}t-F{\"u}nfter Workshop
  der GI-Fachgruppe VR/AR}, 2008.

\bibitem{qian2019make}
Shengju Qian, Kwan-Yee Lin, Wayne Wu, Yangxiaokang Liu, Quan Wang, Fumin Shen,
  Chen Qian, and Ran He.
\newblock Make a face: Towards arbitrary high fidelity face manipulation.
\newblock In {\em Proceedings of the IEEE/CVF International Conference on
  Computer Vision}, pages 10033--10042, 2019.

\bibitem{rothkopf2007task}
Constantin~A Rothkopf, Dana~H Ballard, and Mary~M Hayhoe.
\newblock Task and context determine where you look.
\newblock {\em Journal of vision}, 7(14):16--16, 2007.

\bibitem{shrivastava2017learning}
Ashish Shrivastava, Tomas Pfister, Oncel Tuzel, Joshua Susskind, Wenda Wang,
  and Russell Webb.
\newblock Learning from simulated and unsupervised images through adversarial
  training.
\newblock In {\em Proceedings of the IEEE conference on computer vision and
  pattern recognition}, pages 2107--2116, 2017.

\bibitem{simonyan2014very}
Karen Simonyan and Andrew Zisserman.
\newblock Very deep convolutional networks for large-scale image recognition.
\newblock {\em arXiv preprint arXiv:1409.1556}, 2014.

\bibitem{smith2013gaze}
Brian~A Smith, Qi Yin, Steven~K Feiner, and Shree~K Nayar.
\newblock Gaze locking: passive eye contact detection for human-object
  interaction.
\newblock In {\em Proceedings of the 26th annual ACM symposium on User
  interface software and technology}, pages 271--280, 2013.

\bibitem{sugano2014learning}
Yusuke Sugano, Yasuyuki Matsushita, and Yoichi Sato.
\newblock Learning-by-synthesis for appearance-based 3d gaze estimation.
\newblock In {\em Proceedings of the IEEE Conference on Computer Vision and
  Pattern Recognition}, pages 1821--1828, 2014.

\bibitem{toldo2020unsupervised}
Marco Toldo, Andrea Maracani, Umberto Michieli, and Pietro Zanuttigh.
\newblock Unsupervised domain adaptation in semantic segmentation: a review.
\newblock {\em Technologies}, 8(2):35, 2020.

\bibitem{tzeng2017adversarial}
Eric Tzeng, Judy Hoffman, Kate Saenko, and Trevor Darrell.
\newblock Adversarial discriminative domain adaptation.
\newblock In {\em Proceedings of the IEEE conference on computer vision and
  pattern recognition}, pages 7167--7176, 2017.

\bibitem{ulyanov2016instance}
Dmitry Ulyanov, Andrea Vedaldi, and Victor Lempitsky.
\newblock Instance normalization: The missing ingredient for fast stylization.
\newblock {\em arXiv preprint arXiv:1607.08022}, 2016.

\bibitem{9010675}
Ben Usman, Nick Dufour, Kate Saenko, and Chris Bregler.
\newblock Puppetgan: Cross-domain image manipulation by demonstration.
\newblock In {\em 2019 IEEE/CVF International Conference on Computer Vision
  (ICCV)}, pages 9449--9457, 2019.

\bibitem{wang2022contrastive}
Yaoming Wang, Yangzhou Jiang, Jin Li, Bingbing Ni, Wenrui Dai, Chenglin Li,
  Hongkai Xiong, and Teng Li.
\newblock Contrastive regression for domain adaptation on gaze estimation.
\newblock In {\em Proceedings of the IEEE/CVF Conference on Computer Vision and
  Pattern Recognition}, pages 19376--19385, 2022.

\bibitem{wood2016learning}
Erroll Wood, Tadas Baltru{\v{s}}aitis, Louis-Philippe Morency, Peter Robinson,
  and Andreas Bulling.
\newblock Learning an appearance-based gaze estimator from one million
  synthesised images.
\newblock In {\em Proceedings of the Ninth Biennial ACM Symposium on Eye
  Tracking Research \& Applications}, pages 131--138, 2016.

\bibitem{wood2018gazedirector}
Erroll Wood, Tadas Baltru{\v{s}}aitis, Louis-Philippe Morency, Peter Robinson,
  and Andreas Bulling.
\newblock Gazedirector: Fully articulated eye gaze redirection in video.
\newblock In {\em Computer Graphics Forum}, volume~37, pages 217--225. Wiley
  Online Library, 2018.

\bibitem{wood2015rendering}
Erroll Wood, Tadas Baltrusaitis, Xucong Zhang, Yusuke Sugano, Peter Robinson,
  and Andreas Bulling.
\newblock Rendering of eyes for eye-shape registration and gaze estimation.
\newblock In {\em Proceedings of the IEEE International Conference on Computer
  Vision}, pages 3756--3764, 2015.

\bibitem{worrall2017interpretable}
Daniel~E Worrall, Stephan~J Garbin, Daniyar Turmukhambetov, and Gabriel~J
  Brostow.
\newblock Interpretable transformations with encoder-decoder networks.
\newblock In {\em Proceedings of the IEEE International Conference on Computer
  Vision}, pages 5726--5735, 2017.

\bibitem{xia2020controllable}
Weihao Xia, Yujiu Yang, Jing-Hao Xue, and Wensen Feng.
\newblock Controllable continuous gaze redirection.
\newblock In {\em Proceedings of the 28th ACM International Conference on
  Multimedia}, pages 1782--1790, 2020.

\bibitem{yu2019improving}
Yu Yu, Gang Liu, and Jean-Marc Odobez.
\newblock Improving few-shot user-specific gaze adaptation via gaze redirection
  synthesis.
\newblock In {\em Proceedings of the IEEE/CVF Conference on Computer Vision and
  Pattern Recognition}, pages 11937--11946, 2019.

\bibitem{zakharov2019few}
Egor Zakharov, Aliaksandra Shysheya, Egor Burkov, and Victor Lempitsky.
\newblock Few-shot adversarial learning of realistic neural talking head
  models.
\newblock In {\em Proceedings of the IEEE/CVF International Conference on
  Computer Vision}, pages 9459--9468, 2019.

\bibitem{zhang2018unreasonable}
Richard Zhang, Phillip Isola, Alexei~A Efros, Eli Shechtman, and Oliver Wang.
\newblock The unreasonable effectiveness of deep features as a perceptual
  metric.
\newblock In {\em Proceedings of the IEEE conference on computer vision and
  pattern recognition}, pages 586--595, 2018.

\bibitem{zhang2018revisiting}
Xucong Zhang, Yusuke Sugano, and Andreas Bulling.
\newblock Revisiting data normalization for appearance-based gaze estimation.
\newblock In {\em Proceedings of the 2018 ACM Symposium on Eye Tracking
  Research \& Applications}, pages 1--9, 2018.

\bibitem{zhang2015appearance}
Xucong Zhang, Yusuke Sugano, Mario Fritz, and Andreas Bulling.
\newblock Appearance-based gaze estimation in the wild.
\newblock In {\em Proceedings of the IEEE conference on computer vision and
  pattern recognition}, pages 4511--4520, 2015.

\bibitem{zhang2017s}
Xucong Zhang, Yusuke Sugano, Mario Fritz, and Andreas Bulling.
\newblock It's written all over your face: Full-face appearance-based gaze
  estimation.
\newblock In {\em Proceedings of the IEEE Conference on Computer Vision and
  Pattern Recognition Workshops}, pages 51--60, 2017.

\bibitem{zheng2020self}
Yufeng Zheng, Seonwook Park, Xucong Zhang, Shalini De~Mello, and Otmar
  Hilliges.
\newblock Self-learning transformations for improving gaze and head
  redirection.
\newblock {\em Advances in Neural Information Processing Systems},
  33:13127--13138, 2020.

\bibitem{zou2018unsupervised}
Yang Zou, Zhiding Yu, BVK Kumar, and Jinsong Wang.
\newblock Unsupervised domain adaptation for semantic segmentation via
  class-balanced self-training.
\newblock In {\em Proceedings of the European conference on computer vision
  (ECCV)}, pages 289--305, 2018.

\end{thebibliography}
}

\clearpage

\appendix

\section{Data Pre-processing}
\label{sub:data}
We follow the same data pre-processing pipeline as done in Park \etal ~\cite{park2019few}. The pipeline consists of a normalization technique~\cite{zhang2018revisiting} initially introduced by Sugano \etal~\cite{sugano2014learning}. It is followed by face detection
\cite{hu2017finding} and facial landmarks detection
\cite{deng2018cascade} modules for which open-source implementations are publicly available. The Surrey Face Model~\cite{huber2016multiresolution} is used as a reference 3D face model. Further details can be found in Park \etal~\cite{park2019few}. To summarize, we use the public code\footnote{\url{https://github.com/swook/faze_preprocess}} provided by Park \etal~\cite{park2019few} to produce image patches of size $256\times64$ containing both eyes.

\section{Architecture Details}

\paragraph{Our framework CUDA-GHR.}
We use DenseNet architecture~\cite{huang2017densely} to implement image encoder $\mathbf E_a$. The DenseNet is formed with a growth rate of 32, 4 dense blocks (each with four composite layers), and a compression factor of 1. We use instance
normalization~\cite{ulyanov2016instance} and leaky ReLU activation function ($\alpha$ = 0.01) for all  layers in the network. We remove dropout and $1\times1$ convolution layers. The dimension of latent factor $z^a$ is set to be equal to $16$. Thus, to project CNN features to the latent features, we use global-average pooling and pass through a fully-connected layer to output $16$-dimensional feature from $\mathbf E_a$. The gaze encoder $\mathbf E_g$ is a MLP-based block whose architecture is shown in Table \ref{tab:mlpencoder}. The dimension of $z^g$ is set as $8$. 

For the generator network $\mathbf G$, we use HoloGAN~\cite{nguyen2019hologan} architecture shown in Table \ref{tab:generator}. The latent vector $z$ for each AdaIN~\cite{huang2017arbitrary} input is processed by a 1-layer MLP, and the rotation layer is the same as the one used in the original paper~\cite{nguyen2019hologan}. The latent domain discriminator $\mathbf D_F$ consists of 4 MLP layers as shown in Table \ref{tab:latentdisc}. It takes the input of dimension 24 and gives 1-dimensional output. Both image discriminators $\mathbf D_T$ and $\mathbf D_S$ are PatchGAN~\cite{isola2017image} based networks as used in Zheng \etal~\cite{zheng2020self}. The architecture of the discriminator is described in Table \ref{tab:imagedisc}.

\begin{table}[h]
  \caption{Architecture of gaze encoder $\mathbf E_g$}
  \label{tab:mlpencoder}
  \centering
  \resizebox{\columnwidth}{!}{%
  \begin{tabular}{lcc}
    \hline
    Layer name  & Activation & Output shape \\
    \hline
    Fully connected &  LeakyReLU ($\alpha$ = 0.01) & 2 \\
    Fully connected &  LeakyReLU ($\alpha$ = 0.01) & 2 \\
    Fully connected &  LeakyReLU ($\alpha$ = 0.01) & 2 \\
    Fully connected &  None & 8 \\
    \hline
  \end{tabular}
  }
\end{table}

The task network $\mathcal{T}$ is a ResNet-50~\cite{he2016deep} model with batch normalization~\cite{ioffe2015batch} replaced by instance normalization~\cite{ulyanov2016instance} layers. It takes an input of 256 $\times$ 64 and gives a 4-dimensional output describing pitch and yaw angles for gaze and head directions. It is initialized with ImageNet~\cite{deng2009imagenet} pre-trained weights and is fine-tuned on the GazeCapture training subset for around $190$K iterations. The GazeCapture validation subset is used to select the best-performing model. The initial learning rate is $0.0016$, decayed by a factor of $0.8$ after about $34$K iterations. Adam~\cite{kingma2014adam} optimizer is used for optimization with a weight decay coefficient of $10^{-4}$. The architecture of $\mathcal{T}$ is summarized in Table \ref{tab:tasknet}.

\paragraph{Downstream Tasks.} For gaze and head pose estimation, we use similar architecture as employed for $\mathcal{T}$ shown in Table \ref{tab:tasknet}. For all the experiments, the initial learning rate is $0.0001$ decayed by a factor of $0.5$ after every $1500$ iterations. The pre-trained models are trained for $10$ epochs with a batch size of $64$ while fine-tuning is done for $5$ epochs with a batch size of $32$.

\paragraph{State-of-the-art Baselines.}
We re-implement the ST-ED~\cite{zheng2020self} on images containing both eyes for a fair comparison with our method using the public code\footnote{\url{https://github.com/zhengyuf/STED-gaze}}
available. We use the same hyperparameters as provided by the original implementation. For the accurate comparison, we replaced \textit{tanh} non-linearity with an identity function and removed a constant factor of $0.5\pi$ in all the modules.

\begin{table}[t]
  \caption{Architecture of latent domain discriminator $\mathbf D_F$}
    \label{tab:latentdisc}
  \centering
  \resizebox{\columnwidth}{!}{%
  \begin{tabular}{lcc}
    \hline
    Layer name  & Activation & Output shape \\
    \hline
    Fully connected &  LeakyReLU ($\alpha$ = 0.01) & 24 \\
    Fully connected &  LeakyReLU ($\alpha$ = 0.01) & 24 \\
    Fully connected &  LeakyReLU ($\alpha$ = 0.01) & 24 \\
    Fully connected &  None & 1 \\
    \hline
  \end{tabular}
  }
\end{table}

\begin{table}[t]
  \caption{Architecture of the task network $\mathcal{T}$}
  \label{tab:tasknet}
  \centering
  \resizebox{\columnwidth}{!}{%
  \begin{tabular}{lcc}
    \hline
    Module/Layer name  & Output shape \\
    \hline
    ResNet-50 layers with MaxPool stride=1 &  2048$\times$1$\times$1 \\
    Fully connected & 4 \\
    \hline
  \end{tabular}
  }
\end{table}

\begin{table*}[t]
  \caption{Architecture of the image discriminator networks $\mathbf D_T$ and $\mathbf D_S$. Note that, both the discriminators has the same architecture.}
   \label{tab:imagedisc}
  \centering
  \begin{tabular}{lcccl}
    \hline
    Layer name  & Kernel, Stride, Padding & Activation & Normalization & Output shape \\
    \hline
    Conv2d & 4$\times$4, 2, 1 & LeakyReLU ($\alpha$ = 0.2) & - & 64$\times$ 32$\times$128 \\
    Conv2d & 4$\times$4, 2, 1 & LeakyReLU ($\alpha$ = 0.2) & InstanceNorm & 128$\times$16$\times$64 \\
    Conv2d & 4$\times$4, 2, 1 & LeakyReLU ($\alpha$ = 0.2) & InstanceNorm & 256$\times$8$\times$32 \\
    Conv2d & 4$\times$4, 1, 1 & LeakyReLU ($\alpha$ = 0.2) & InstanceNorm & 512$\times$7$\times$31 \\
    Conv2d & 4$\times$4, 1, 1 & - & - & 1$\times$6$\times$30 \\    
    \hline
  \end{tabular}
\end{table*}

\begin{table*}[t]
  \caption{Architecture of the generator network $\mathbf G$}
   \label{tab:generator}
  \centering
  \begin{tabular}{lcccl}
    \hline
    Layer name  & Kernel & Activation & Normalization & Output shape \\
    \hline
    Learned Constant Input & - & - & - & 512$\times$4$\times$2$\times$8 \\
    Upsampling & - & - & - & 512$\times$8$\times$4$\times$16 \\
    
    Conv3d & 3$\times$3$\times$3 & LeakyReLU & AdaIN & 256$\times$8$\times$ 4$\times$16 \\
    
    Upsampling & - & - & - & 256$\times$16$\times$8$\times$32 \\
    
    Conv3d & 3$\times$3$\times$3 & LeakyReLU & AdaIN & 128$\times$16$\times$8$\times$32 \\
    
    Volume Rotation & - & - & - & 128$\times$16$\times$8$\times$32 \\
    
    Conv3d & 3$\times$3$\times$3 & LeakyReLU & - & 64$\times$16$\times$8$\times$32 \\
    
    Conv3d & 3$\times$3$\times$3 & LeakyReLU & - & 64$\times$16$\times$8$\times$32 \\
    
    Reshape & - & - & - & (64 $\cdot$ 16)$\times$8$\times$32 \\
    
    Conv2d & 1$\times$1 & LeakyReLU & - & 512$\times$8$\times$32 \\
    
    Conv2d & 4$\times$4 & LeakyReLU & AdaIN & 256$\times$8$\times$32 \\
    Upsampling & - & - & - & 256$\times$16$\times$32 \\

    Conv2d & 4$\times$4 & LeakyReLU & AdaIN & 64$\times$16$\times$64 \\
    Upsampling & - & - & - & 64$\times$32$\times$128 \\
    
    Conv2d & 4$\times$4 & LeakyReLU & AdaIN & 32$\times$32$\times$128 \\
    Upsampling & - & - & - & 32$\times$64$\times$256 \\
    
    Conv2d & 4$\times$4 & Tanh & - & 3$\times$64$\times$256 \\

    \hline
  \end{tabular}
\end{table*}

\section{Additional Results}
\label{addnres}


In Figures \ref{fig:overalladdmpii} and  \ref{fig:overalladdcol}, we show additional qualitative results for both target datasets, namely, MPIIGaze and Columbia. Figure \ref{fig:gazeswapped_1} and \ref{fig:gazeswapped_2} represent gaze redirected images and Figure \ref{fig:headswapped_1} and \ref{fig:headswapped_2} show head redirected images.

\begin{figure*}[t!]
    \centering
    \begin{subfigure}[h]{\linewidth}
        \includegraphics[width=\linewidth]{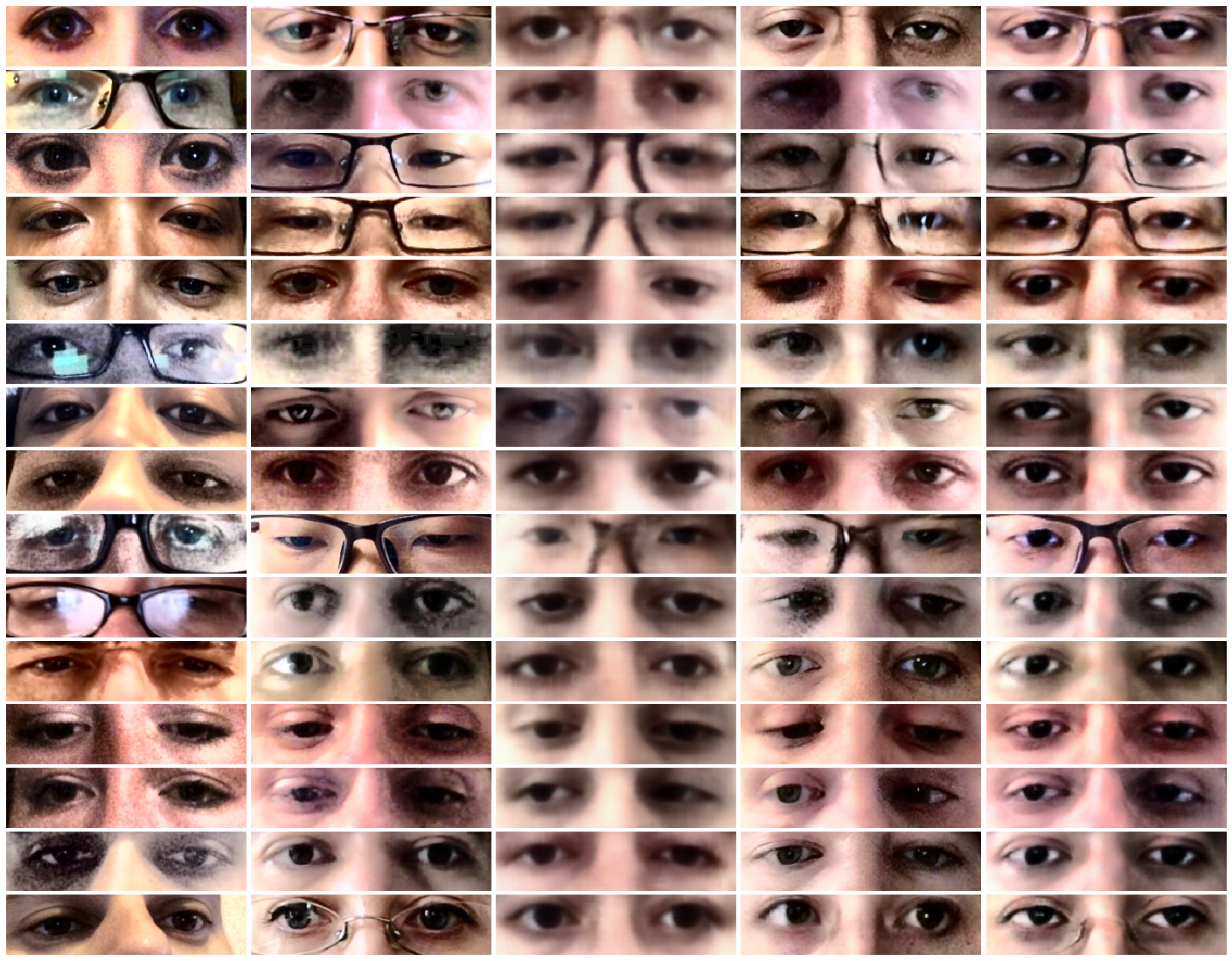}
        \caption*{\scriptsize{Gaze Source \hspace{2.5cm} Input Image  \hspace{2.5cm} FAZE  \hspace{2.5cm} ST-ED  \hspace{2.5cm}  CUDA-GHR}}
        \vspace{0.2cm}
       \caption{Gaze Redirected images for MPIIGaze dataset (\textit{GazeCapture$\rightarrow$MPIIGaze})}
        \label{fig:gazeswapped_1}
    \end{subfigure}
\end{figure*}
\begin{figure*}[ht]\ContinuedFloat
	\begin{subfigure}[h]{\linewidth}
    \centering
        \includegraphics[width=1\linewidth]{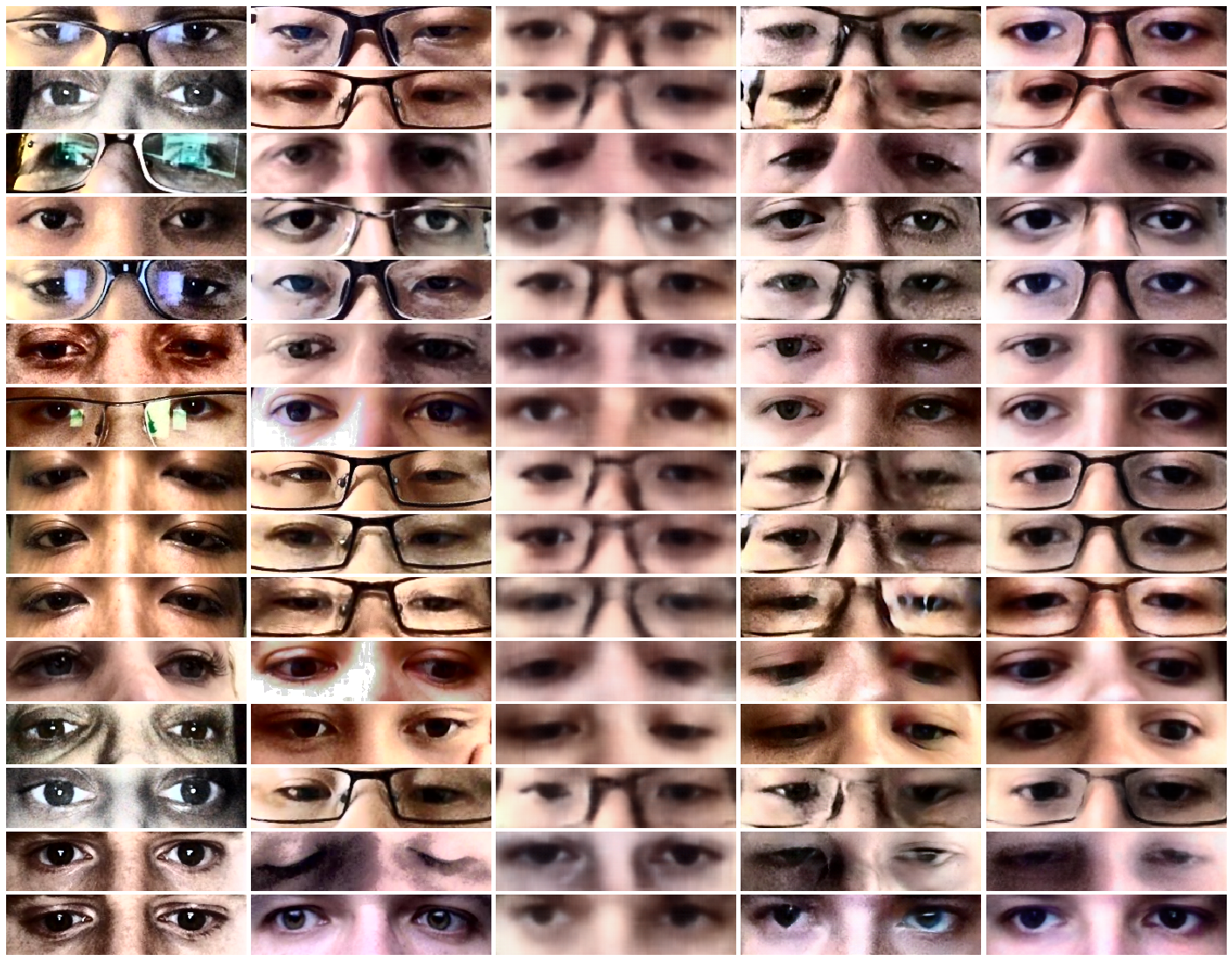}
        \caption*{\scriptsize{Head Source \hspace{2.5cm} Input Image  \hspace{2.5cm} FAZE  \hspace{2.5cm} ST-ED  \hspace{2.5cm}  CUDA-GHR}}
        \vspace{0.2cm}
        \caption{Head Redirected images for MPIIGaze dataset (\textit{GazeCapture$\rightarrow$MPIIGaze)}}
        \label{fig:headswapped_1}
    \end{subfigure}
\caption{\textbf{Additional Qualitative Results ({\textit{GazeCapture$\rightarrow$MPIIGaze}):}} More qualitative results on the MPIIGaze dataset. \ref{fig:gazeswapped_1} shows the gaze redirected images and \ref{fig:headswapped_1} shows the head redirected images. The first column shows the gaze/head pose source image from which gaze/head pose information is used to redirect. The second column shows the input image from the MPIIGaze  dataset. Best viewed in color.}
\label{fig:overalladdmpii}
\end{figure*}

\begin{figure*}[t!]
    \centering
    \begin{subfigure}[h]{\linewidth}
        \includegraphics[width=\linewidth]{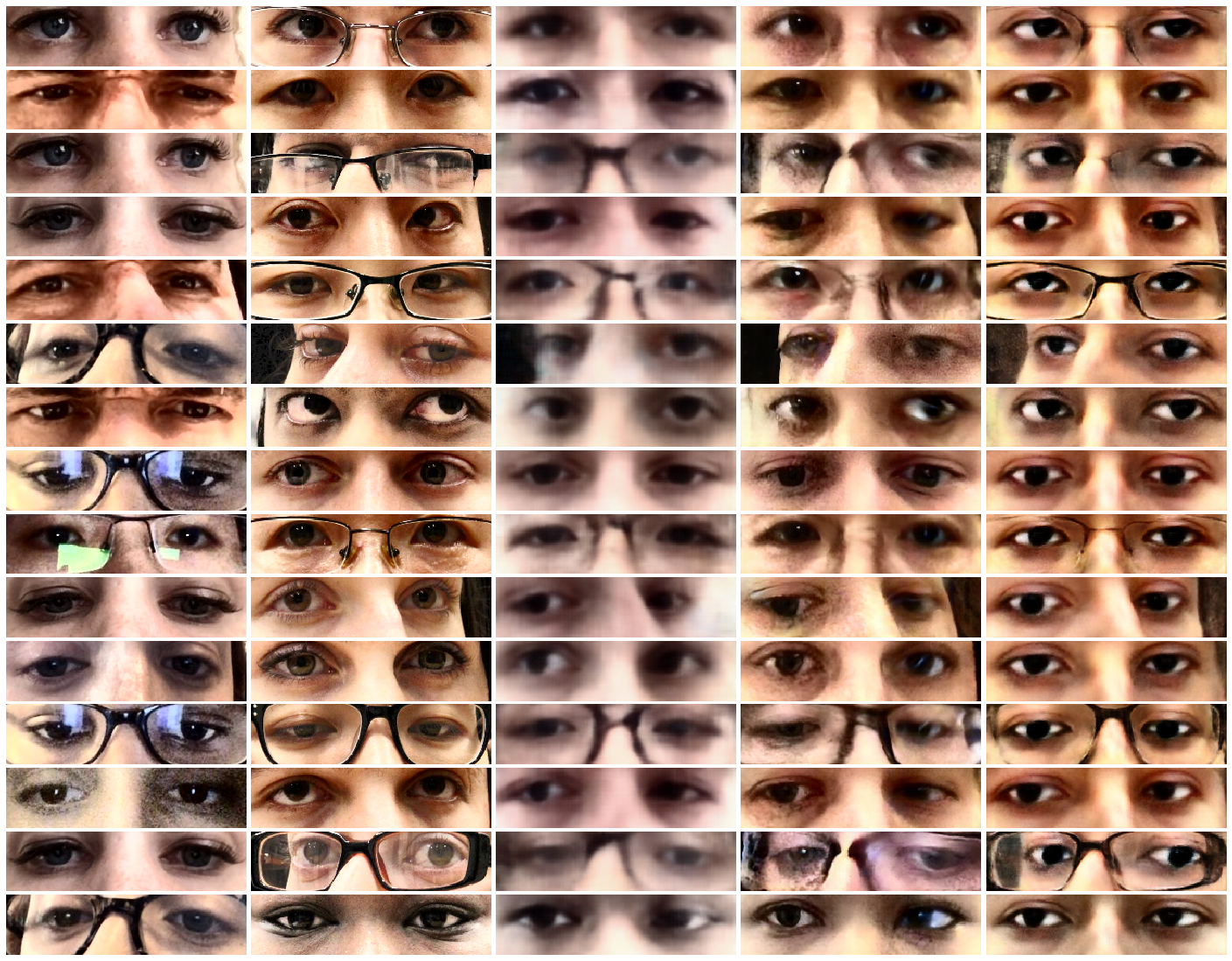}
        \caption*{\scriptsize{Gaze Source \hspace{2.5cm} Input Image  \hspace{2.5cm} FAZE  \hspace{2.5cm} ST-ED  \hspace{2.5cm}  CUDA-GHR}}
        \vspace{0.2cm}
        \caption{Gaze Redirected images for Columbia dataset (\textit{GazeCapture$\rightarrow$Columbia})}
    \label{fig:gazeswapped_2}
    \end{subfigure}
\end{figure*}
\begin{figure*}[ht]\ContinuedFloat
	\begin{subfigure}[h]{\linewidth}
    \centering
        \includegraphics[width=1\linewidth]{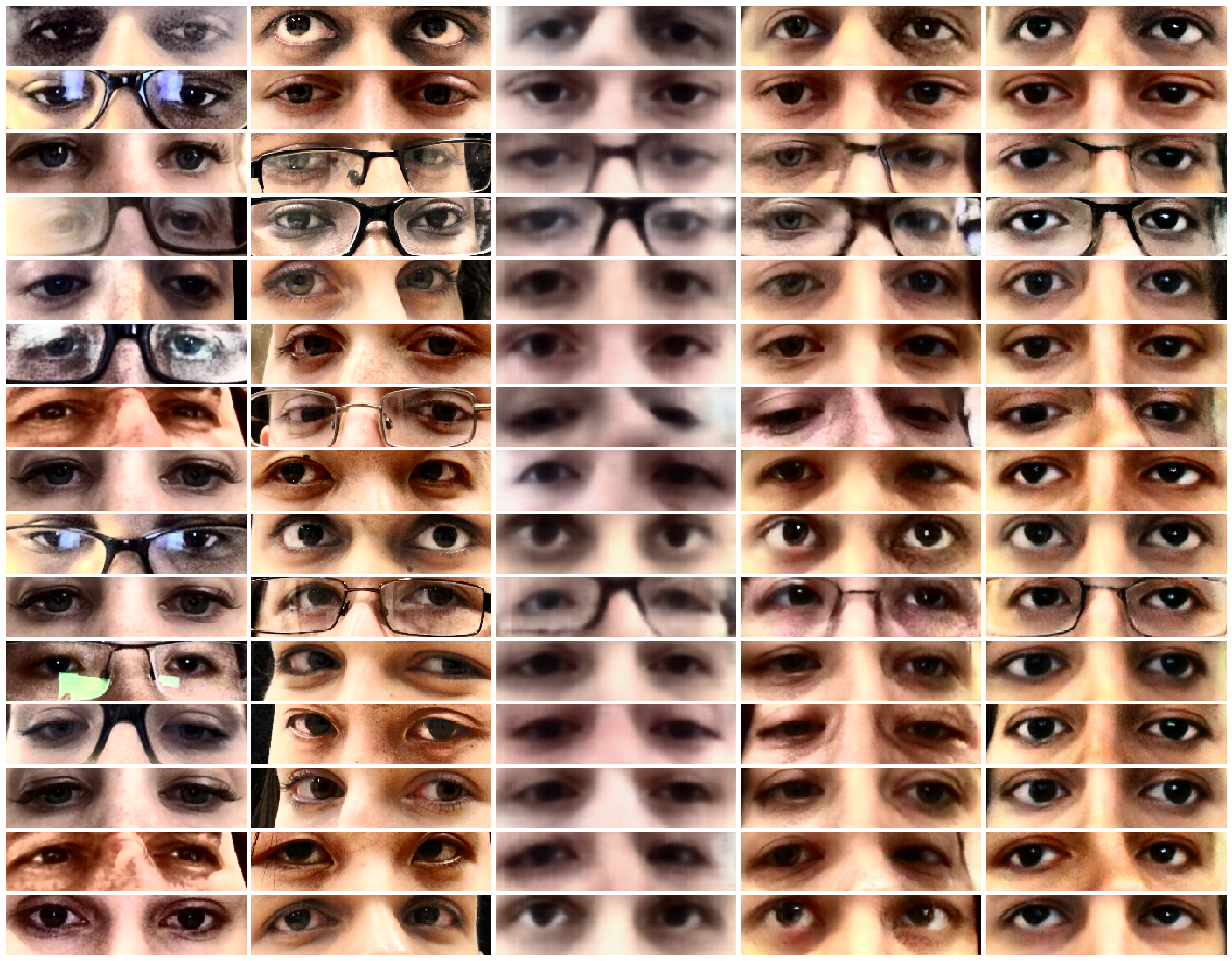}
        \caption*{\scriptsize{Head Source \hspace{2.5cm} Input Image  \hspace{2.5cm} FAZE  \hspace{2.5cm} ST-ED  \hspace{2.5cm}  CUDA-GHR}}
        \vspace{0.2cm}
        \caption{Head Redirected images for Columbia dataset (\textit{GazeCapture$\rightarrow$Columbia)}}
        \label{fig:headswapped_2}
    \end{subfigure}
\caption{\textbf{Additional Qualitative Results (\textit{GazeCapture$\rightarrow$Columbia}):} Qualitative results on the Columbia dataset. \ref{fig:gazeswapped_2} shows the gaze redirected images and \ref{fig:headswapped_2} shows the head redirected images. The first column shows the gaze/head pose source image from which gaze/head pose information is used to redirect. The second column shows the input image from the Columbia  dataset. Best viewed in color.}
\label{fig:overalladdcol}
\end{figure*}

\end{document}